\documentclass[letterpaper]{article} 
\usepackage[preprint]{aaai2027}  
\usepackage[hyphens]{url}  
\usepackage{graphicx} 
\usepackage{natbib}  
\usepackage{caption} 
\usepackage{algorithm}
\usepackage{algpseudocode}
\usepackage{amsmath, amssymb}
\usepackage{multirow}
\usepackage{booktabs}
\usepackage[table]{xcolor}          
\definecolor{ourshl}{RGB}{237,233,250}   
\usepackage{graphicx}
\usepackage{trimclip}      
\usepackage{calc}
\usepackage{xcolor}
\usepackage{caption}
\usepackage{adjustbox} 
  \usepackage{array}
  \usepackage{tabularx}
  \usepackage{booktabs}
  \usepackage{subcaption}
\usepackage{newfloat}
\usepackage{listings}
\DeclareCaptionStyle{ruled}{labelfont=normalfont,labelsep=colon,strut=off} 
\floatstyle{ruled}
\newfloat{listing}{tb}{lst}{}
\floatname{listing}{Listing}

\usepackage{booktabs}

\title{PROVE: Training-Free Prompt Recovery using Verifiable Evidence}

\author {
    Rupayan Mallick,
    Mahsa Khoshnoodi,
    Sarah Adel Bargal
}
\affiliations {
    Department of Computer Science, Georgetown University\\
}

\usepackage{bibentry}

\begin{document}

\maketitle

 \begin{abstract}

Modern text-to-image models can generate highly realistic images from natural-language prompts, while recent advances in prompt inversion have made it increasingly feasible to recover those prompts from generated outputs, raising new concerns for copyright protection and content ownership. As prompt marketplaces emerge, recovered prompts can enable both the unauthorized reproduction and redistribution of copyrighted creative works, and the exposure of the prompts that encode an artist's creative recipe in AI-generated content. Existing prompt inversion methods rely on gradient-based optimization, autoregressive captioning, or reinforcement learning. However, optimization-based methods often produce unreadable prompts, captioning methods hallucinate unverified details, and RL-based approaches frequently overfit to specific generators while introducing evaluation circularity. We introduce \emph{PROVE} (Prompt Recovery with Verified Evidence), a training-free, black-box prompt inversion attack that reconstructs prompts by composing verifiable scene descriptions rather than optimizing token sequences, targeting both original copyrighted works and AI-generated content. The resulting prompts are fully auditable, with every recovered claim grounded in explicit image evidence, and are formalized through a precision-constrained recall maximization objective. Across MS-COCO, Flickr30K, and Lexica, using state-of-the-art text-to-image generators, \emph{PROVE} consistently outperforms optimization, captioning, and RL-based baselines on image similarity (DINO, LPIPS) and text-image alignment (CLIP), without any training, generator access, or fine-tuning, demonstrating a stronger and more practical prompt inversion attack.

\end{abstract}

\section{Introduction}

Text-to-image generation has been transformed by diffusion- and flow-based models such as Stable Diffusion \cite{esser2024scaling}, DALL-E \cite{ramesh2021zero}, FLUX \cite{labs2025flux1kontextflowmatching}, and Imagen \cite{saharia2022photorealistic}. These models can synthesize highly realistic images from natural-language descriptions, with image quality and fidelity largely determined by their ability to interpret user prompts. This capability has democratized image creation, editing, and content generation, enabling non-experts to produce sophisticated visual content through simple textual instructions. However, the same accessibility introduces significant security, safety, and intellectual property concerns. Prompt manipulation and adversarial prompting can facilitate harmful generation, including the creation of copyrighted derivatives and other forms of misuse~\cite{bariach2024towards}.

\definecolor{proveGreen}{RGB}{34,139,34}
\begin{figure}[t]
\centering
\setlength{\tabcolsep}{1pt}
\begin{tabular}{@{}cccc@{}}
\multicolumn{2}{c}{\scriptsize\textbf{MS-COCO}} &
\multicolumn{2}{c}{\scriptsize\textbf{Lexica}} \\[1pt]
\tiny Original & \tiny\textbf{\textcolor{proveGreen}{PROVE}} &
\tiny Original & \tiny\textbf{\textcolor{proveGreen}{PROVE}} \\[1pt]
\includegraphics[width=0.24\linewidth,height=0.24\linewidth]{used_photos_inversion/mscoco_Original.jpg} &
\includegraphics[width=0.24\linewidth,height=0.24\linewidth]{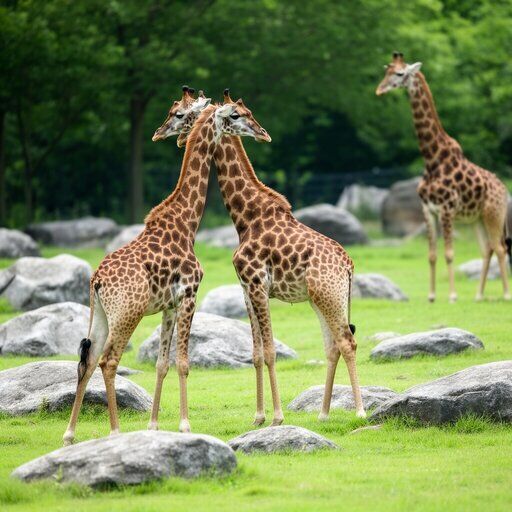} &
\includegraphics[width=0.24\linewidth,height=0.24\linewidth]{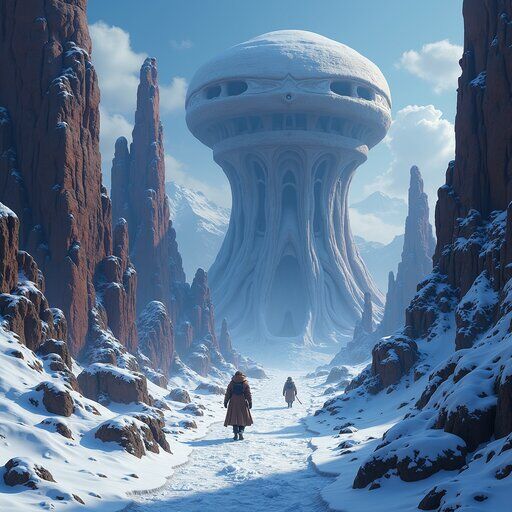} &
\includegraphics[width=0.24\linewidth,height=0.24\linewidth]{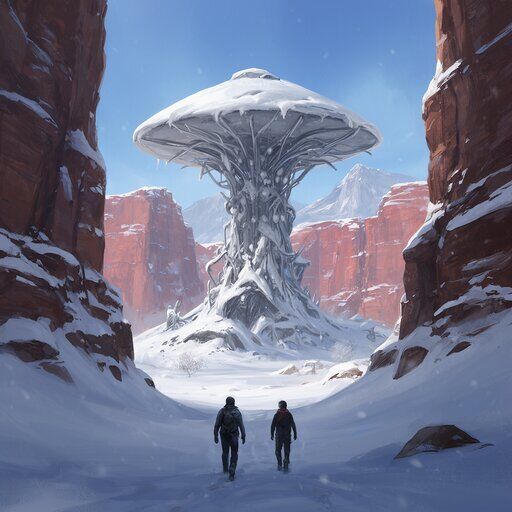} \\
\end{tabular}
\caption{PROVE regenerations of a real MS-COCO image and an AI-generated image from a Lexica prompt.}
\label{fig:orig-prove-1row}
\end{figure}

The emergence of prompt marketplaces, where carefully engineered prompts are shared or sold~\cite{shen2024prompt,10.1145/3731715.3733309}, further amplifies these risks for both original non-AI content, and AI-generated content. For original copyrighted content, prompt inversion can enable unauthorized reproduction. For AI-generated content, prompt inversion, also called prompt recovery or prompt stealing, can result in prompts that encode valuable creative knowledge and generation recipes. Some AI artists have refused to share their prompts see~\cite{cpr2023allen}. For both original and AI-generated content, the risk is recproducing the target image, or elements of a target image, as closely as possible.

Beyond security concerns, prompt inversion is important for understanding model behavior, studying text-image alignment, and improving model transparency. Recovering prompts from generated images can support model provenance, attribution, and interpretable AI by exposing the semantic factors underlying image generation. Existing approaches span both white-box settings, where model parameters or gradients are available, and black-box settings, where access is restricted to APIs or generated outputs. Understanding how effectively prompts can be reconstructed in these settings is therefore critical for both advancing trustworthy AI and assessing emerging risks to intellectual property and content ownership.

\begin{figure*}[t]
    \centering
    \includegraphics[width=1.0\linewidth]{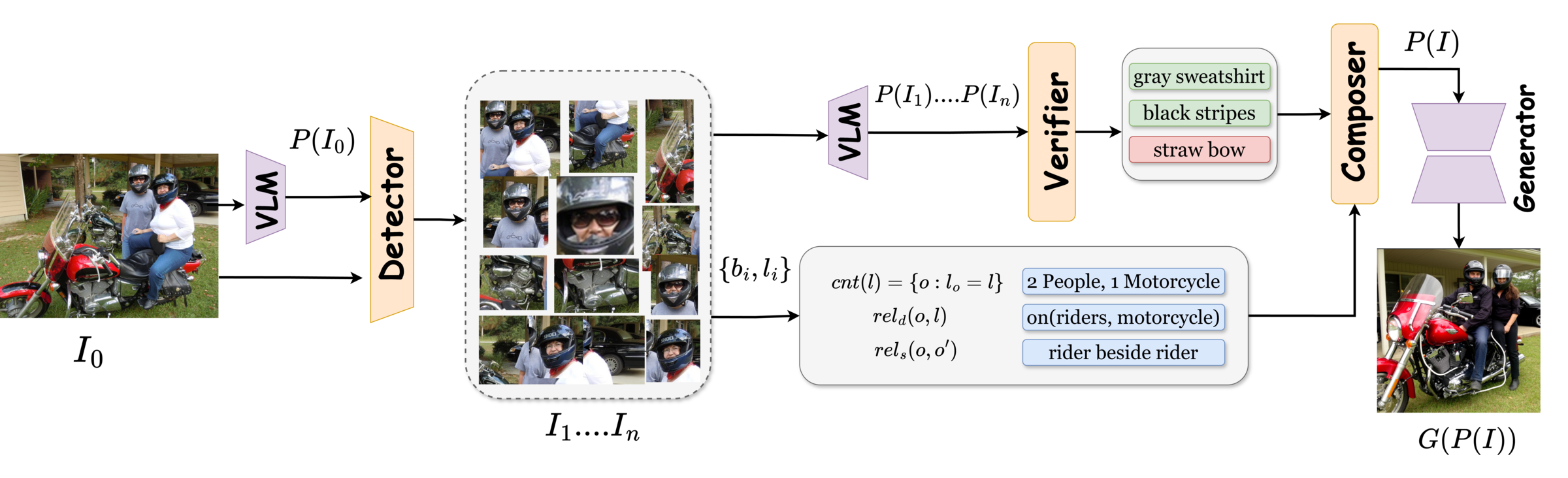}
    \caption{\textbf{Overview of the proposed framework.} Given an image $I_0$, a frozen VLM $g_{\phi}$ and a text-based object detector localize regions
$I_1,\dots,I_n$. Per-region claims from the VLM are filtered by an evidence-gated \emph{Verifier}, while counts and spatial relations are derived
deterministically in the \emph{geometric extraction} block (gray). The \emph{Composer} serializes the surviving claims into the final prompt $P(I)$, which the generator renders as $G(P(I))$.}
    \label{fig:placeholder}
\end{figure*}

In white-box settings, 
prompt inversion is typically formulated as an optimization problem in either the text-embedding space or the diffusion latent space. These methods leverage the alignment between vision and language encoders to iteratively refine prompts toward a target image. Representative examples include PEZ~\cite{wen2023hard}, PH2P~\cite{mahajan2024prompting}, and STEPs~\cite{Qiu2025STEPSSP} -all optimization-based approaches. While effective at improving image similarity, optimization-based methods often produce prompts that drift away from natural language, yielding token sequences that are difficult for humans to interpret. Moreover, the discrete nature of language introduces a fundamental challenge, as textual prompts are inherently incompatible with conventional continuous optimization techniques.

In black-box settings, there is no knowledge of model parameters and access is limited to APIs or generated images, prompt inversion is commonly approached through image captioning and vision-language models (VLMs). Methods based on BLIP~\cite{li2022blip}, CLIP-IG~\cite{clip_interrogator}, and VLMs generate prompts directly from visual content by leveraging pretrained vision and language encoders. Although these approaches produce fluent and interpretable descriptions, the prompts aren't optimized, and often lack mechanisms for verifying whether generated claims are actually supported by the image, reducing prompt fidelity.

More recently, PromptStealer~\cite{shen2024prompt} decomposed prompt stealing into subject and modifier recovery, using a fine-tuned captioning model to infer the subject and a multi-label detector to predict stylistic modifiers, while PromptMiner~\cite{li2025promptminer} proposed an RL-based framework that mines prompts from online marketplaces and learns to recover them through reinforcement learning, followed by fuzz-based optimization of stylistic modifiers. While promising, this approach relies on the same image generation model for both target and reconstructed images and use the same objective function for the RL optimization and metric computation, raising concerns about generator-specific overfitting and evaluation circularity. 

We propose a state-of-the-art black-box prompt inversion attack, \emph{PROVE}, that is evidence-based, training-free, and not generator-specific. \emph{PROVE} consists of four stages: grounded claim proposal using frozen vision-language models and open-vocabulary detection; evidence-gated verification, where each atomic claim must satisfy a label-anchored CLIP margin on its corresponding image region; geometric fact extraction, which deterministically derives counts and spatial relationships; and deterministic composition, which serializes verified claims into a fluent prompt.We show that \emph{PROVE's} generated prompts are SOTA on MS-COCO, Flickr30K, and Lexica (Figure~\ref{fig:orig-prove-1row}), and that the proposed attack is robust to reasonable defenses.

\section{Related Work}
\label{sec:related}

\paragraph{Prompt inversion for text-to-image models.}
Prompt inversion recovers a textual description from a target image such that regenerating from that text reproduces the image. Two families exist. Soft inversion optimizes continuous pseudo-tokens that are appended to the model vocabulary, as in Textual Inversion \cite{gal2022image} and multi-concept customization \cite{kumari2023multi}; the recovered artifact is an embedding and is not human readable. Hard inversion instead returns discrete natural
language. PEZ performs gradient-based optimization over discrete tokens to recover prompts that reproduce a target image \cite{wen2023hard}. PH2P optimizes interpretable prompts directly from a target image through a delayed
projection scheme and L-BFGS \cite{mahajan2024prompting} EDITOR initializes from a captioning model, refines in latent space, and decodes back to text
\cite{li2025editor}. VGD removes the need for gradients, pairing a language model with CLIP guidance
to produce coherent, human-readable prompts \cite{kim2025visually}. We additionally evaluate STEPS and PromptMiner \cite{Qiu2025STEPSSP,li2025promptminer}. Captioning models such as BLIP-2 are commonly used as a fast approximate baseline for this task
\cite{williams2024prompt}. We note that this line of work is distinct from latent inversion methods such as null-text and negative-prompt inversion
\cite{mokady2023null,miyake2025negative}, which recover a noise trajectory for editing rather than a textual artifact, therefore falls outside the threat model we consider.


All of the hard inversion methods we evaluate take a source image as input and
return text. They were developed for prompt engineering, data attribution,
model provenance, and watermark validation, and their reported evaluations
measure CLIP-space image and text similarity, distributional alignment between
generations from the original and recovered prompts, and human judgments of
interpretability \cite{wen2023hard,mahajan2024prompting,williams2024prompt}.
We repurpose them as measurement instruments and evaluate them against a
criterion their original benchmarks do not report.

\paragraph{Inversion as a privacy threat.}
A smaller body of work already treats inversion adversarially. Prompt stealing attacks such as PromptStealer reconstruct the prompt behind a generated image in order to appropriate its style or content \cite{shen2024prompt}, and Ye et
al. extend this threat across modalities with a unified cross-modal inversion approach \cite{309620}. A parallel line of work targets the training data rather than the prompt: membership inference attacks determine whether an image was used in training \cite{wu2022membership}, extraction and replication studies show that diffusion models memorize and regurgitate individual training
examples \cite{carlini2023extracting,somepalli2023diffusion}, and fine-tuning
has been shown to amplify this risk \cite{li2024shake}. ML-Doctor provides a
complementary, holistic risk assessment of inference attacks against machine
learning models \cite{liu2022ml}. Across this literature the asset at risk is
either the prompt or the training set, and the adversary is assumed to hold
some privileged relationship to the model. Our threat model assumes neither.

\section{Methods}
\label{subsec:method_overview}

\begin{algorithm}[t]
\caption{Grounded, verified prompt inversion (training-free, no generator in
the loop)}
\label{alg:pipeline}
\begin{algorithmic}[1]
\Require image $I$; VLM $g_\phi$; detector $D_\theta$; CLIP $f$;
         margin $\delta$; crop padding $p$
\Ensure  prompt pair $(c_g, P(I))$; verified hierarchy $H(I)$
\Statex {\footnotesize\textit{Notation:} $\kappa(b_i,b_k)$: fraction of $b_i$
        inside $b_k$; $\nu_i(\mathcal{S})\!=\!\arg\min_{k\in\mathcal{S}}
        \lVert\mathrm{cen}(b_i)\!-\!\mathrm{cen}(b_k)\rVert$ (nearest center);
        $\Pi_i(\tau,r)\!=\!\{k:\kappa(b_i,b_k)\!\ge\!\tau,\,|b_i|\!\le\!r|b_k|\}$.}
\smallskip
\State $c_g \gets g_\phi(I, \pi_{cap})$
       \Comment{(a) global caption}
\State $\mathcal{A} \gets \{a \in S(g_\phi(I, \pi_{scene})) :
        E(a; I, \texttt{photo}) \ge -\delta\}$
       \Comment{verified scene attributes}
\State $Q \gets N(c_g) \cup N\big(g_\phi(I, \pi_{inv})\big)$
       \Comment{query set = recall ceiling}
\State $\{(b_i, \ell_i)\} \gets \mathrm{NMS}\big(D_\theta(I_{512}, Q)\big)$
       \Comment{detect on working copy}
\ForAll{regions $i$}
  \State $\alpha(o_i) \gets
    \nu_i\big(\Pi_i(\tau_a,1)\big)$ if $\ell_i\!\in\!\mathcal{L}_{att}$;\;
    $\mathrm{scene}$ if $\ell_i\!\in\!\mathcal{L}_{scn}$;\;
    $\nu_i\big(\Pi_i(\max(\tau_a,\lambda),\gamma)\big)$ otherwise
  \Statex \hfill\Comment{priors, then containment geometry}
\EndFor
\ForAll{regions $i$}
    \State $x_i \gets \mathcal{C}_p(I, \rho(b_i))$
        \Comment{native-resolution crop}
    \State $d_i \gets g_\phi(x_i, \pi_{slot} \mid \ell_i)$
        \Comment{slot-schema description}
    \State $K_i \gets \{\, c \in S(d_i) :
            E(c; x_i, \ell_i) \ge -\delta \,\}$
        \Comment{(b) evidence gate; salvage; $K_i \gets \{\ell_i\}$ if empty}
\EndFor
\State $\mathcal{F}_{geo} \gets$ counts $n_\ell$, support $\mathrm{on}(o,u)$,
       relations $\mathrm{rel}(o,o')$ from $\{(b_i, \ell_i)\}$
       \Comment{(c) detector-exact; bypass $E$}
\State $H(I) \gets \big\langle\, c_g \oplus \mathcal{A} \oplus
       \mathcal{F}_{geo},\; \{(\ell_k, b_k, K_k)\}_k \,\big\rangle$
\State $P(I) \gets \mu\big(H(I)\big)$
       \Comment{(d) deterministic composition; no truncation}
\State \Return $(c_g,\, P(I)),\; H(I)$
\end{algorithmic}
\end{algorithm}

\newlength{\cellside}\setlength{\cellside}{2.05cm} 
\setlength{\tabcolsep}{1.6pt}                      
\newlength{\cellpad}\setlength{\cellpad}{\tabcolsep}   
\newlength{\rowgap}\setlength{\rowgap}{2pt}        
\newcommand{\vc}[1]{\adjustbox{valign=c,margin=0pt \cellpad}{#1}}
\renewcommand{\arraystretch}{1}
\newlength{\labwidth}\setlength{\labwidth}{0.6cm}  
\newlength{\figwidth}
\setlength{\figwidth}{\labwidth}
\addtolength{\figwidth}{9\cellside}
\addtolength{\figwidth}{18\tabcolsep}
\addtolength{\figwidth}{4pt}
\definecolor{provegreen}{RGB}{0,150,60}
\newlength{\imw}\newlength{\imh}\newlength{\croplen}
\newcommand{\sq}[1]{%
  \settowidth{\imw}{\includegraphics{#1}}%
  \settoheight{\imh}{\includegraphics{#1}}%
  \ifdim\imw>\imh
    \settowidth{\imw}{\includegraphics[height=\cellside]{#1}}%
    \setlength{\croplen}{\dimexpr(\imw-\cellside)/2\relax}%
    \ifdim\croplen<0pt\setlength{\croplen}{0pt}\fi
    \edef\trimspec{\the\croplen\space 0pt \the\croplen\space 0pt}%
    \expandafter\clipbox\expandafter{\trimspec}{\includegraphics[height=\cellside]{#1}}%
  \else
    \settoheight{\imh}{\includegraphics[width=\cellside]{#1}}%
    \setlength{\croplen}{\dimexpr(\imh-\cellside)/2\relax}%
    \ifdim\croplen<0pt\setlength{\croplen}{0pt}\fi
    \edef\trimspec{0pt \the\croplen\space 0pt \the\croplen}%
    \expandafter\clipbox\expandafter{\trimspec}{\includegraphics[width=\cellside]{#1}}%
  \fi
}
\newcommand{\dslabel}[1]{%
  \vc{\rotatebox[origin=c]{90}{\rmfamily\bfseries\itshape\fontsize{11}{13}\selectfont #1}}%
}
\newcommand{\gridrow}[3]{%
  \dslabel{#1} &
  \vc{\sq{#2/Original.#3}} & \vc{\sq{#2/PROVE.jpg}} & \vc{\sq{#2/PromptMiner.jpg}} &
  \vc{\sq{#2/STEPs.jpg}} & \vc{\sq{#2/VGD.jpg}} & \vc{\sq{#2/PH2P.jpg}} &
  \vc{\sq{#2/VLM-as-Expert.jpg}} & \vc{\sq{#2/BLIP.jpg}} & \vc{\sq{#2/CLIP-IG.jpg}}\\[\rowgap]%
}
\newcommand{\colhead}{%
  \small
  & Original & {\bfseries\color{provegreen}PROVE} & PromptMiner & STEPs & VGD &
  PH2P & VLM-Expert & BLIP & CLIP-IG\\[3pt]%
}

\newcommand{\qualgrid}{%
  \begin{minipage}{\figwidth}
    \begin{tabular}{@{}c*{9}{c}@{}}
      \colhead
      \gridrow{}{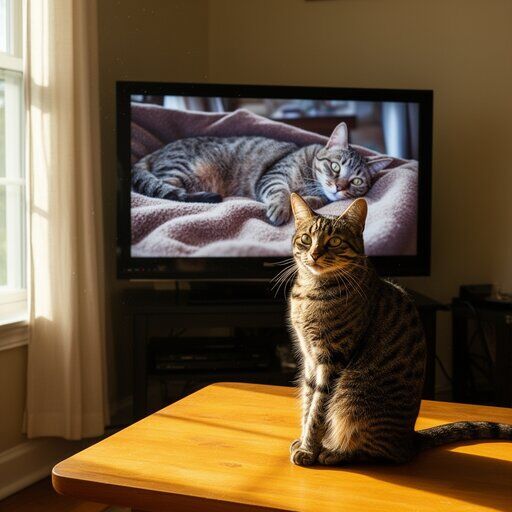}{jpg}
      \gridrow{MS-COCO}{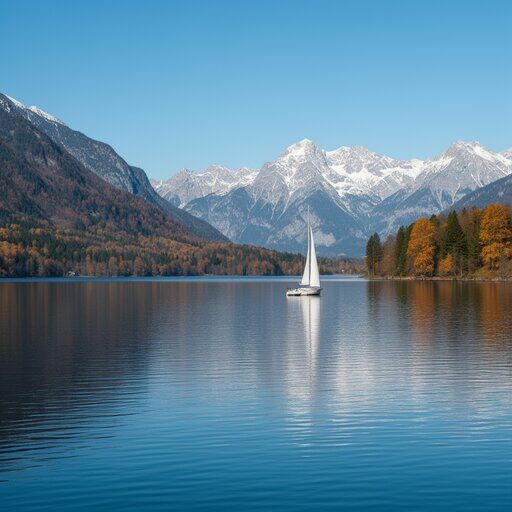}{jpg}
      \gridrow{}{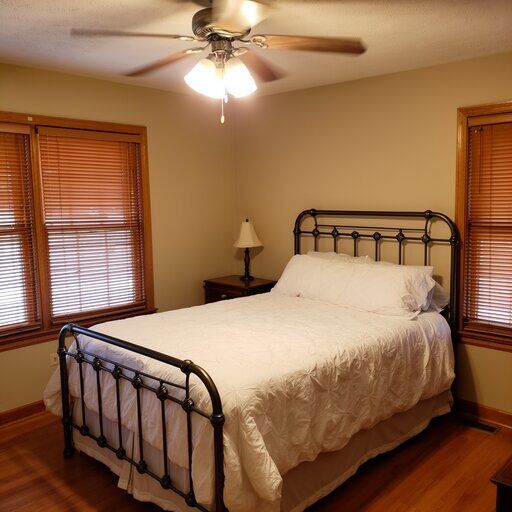}{jpg}
      \noalign{\vskip 7pt}
      \gridrow{}{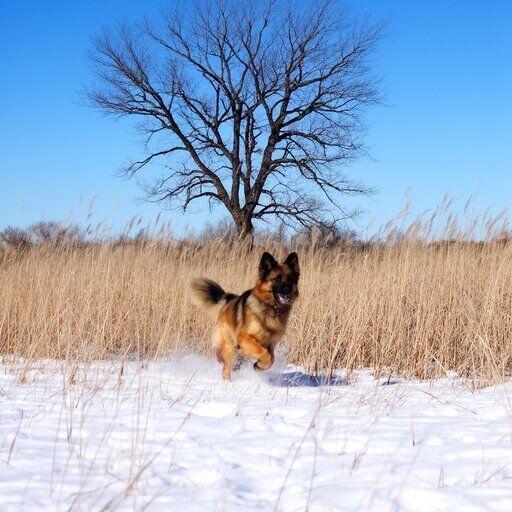}{jpg}
      \gridrow{Flickr}{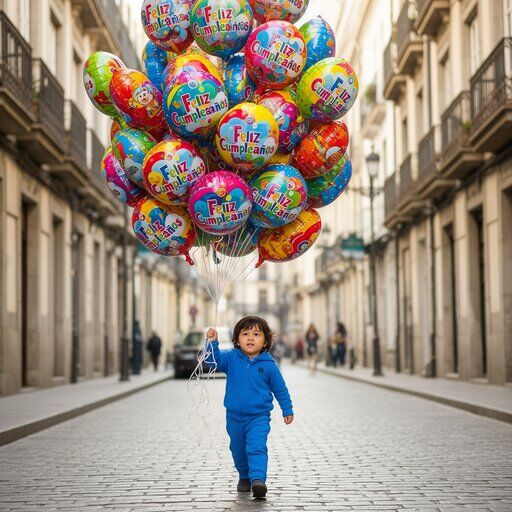}{jpg}
      \gridrow{}{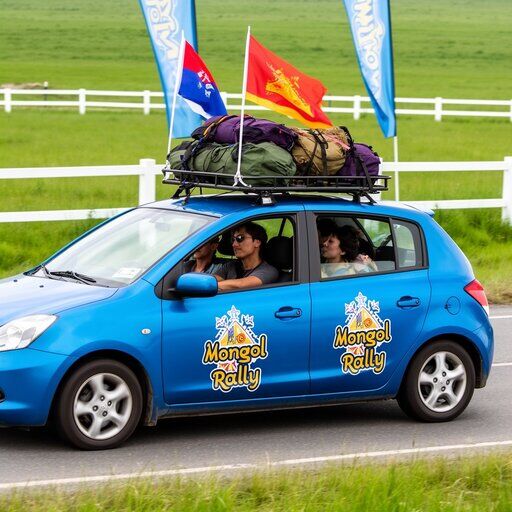}{jpg}
      \noalign{\vskip 7pt}
      \gridrow{}{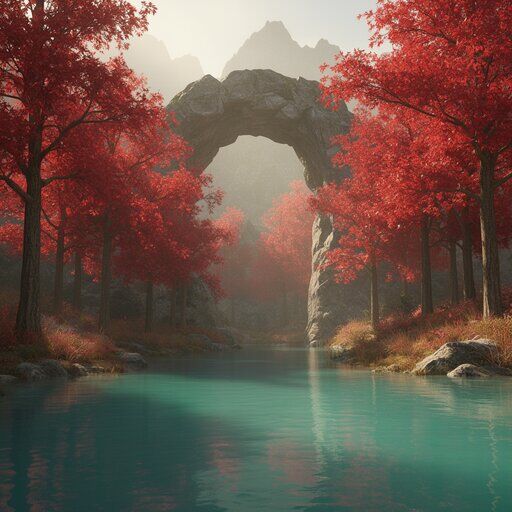}{jpg}
      \gridrow{Lexica}{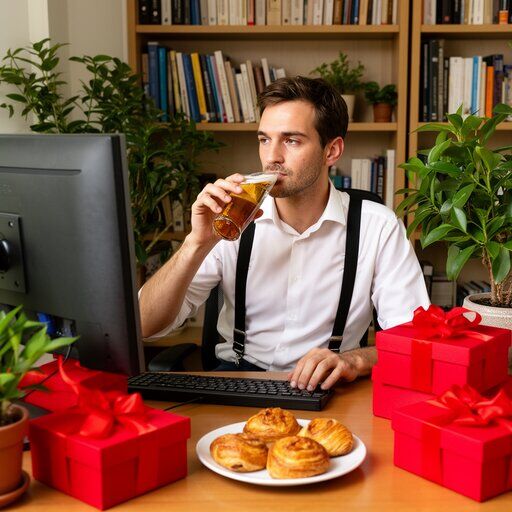}{jpg}
      \gridrow{}{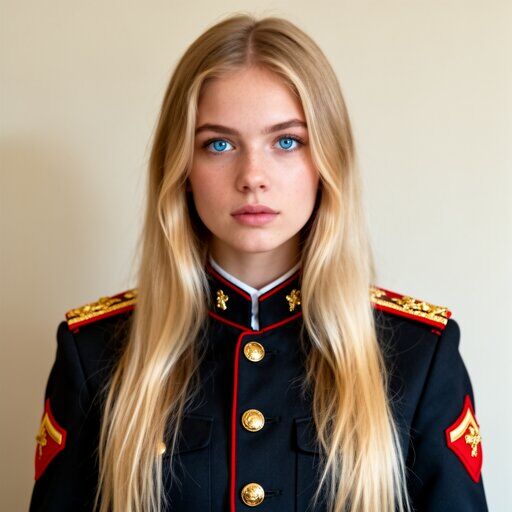}{jpg}
    \end{tabular}
  \end{minipage}%
}

\begin{figure*}[tp]
  \centering
  \adjustbox{max width=\textwidth,
             max totalheight=0.88\textheight}{\qualgrid}
  \caption{Qualitative comparison across MS-COCO, Flickr30K, and Lexica.
    For each example the original image is shown alongside reconstructions
    from PROVE (ours) and the competing methods.}
  \label{fig:qual}
\end{figure*}

The task of recovering the prompt used to generate an image is referred to in the literature as either \textit{prompt inversion} or \textit{prompt stealing},in both cases the objective is to produce a prompt that reproduces the target image. Existing methods also differ in their assumptions about model access~\cite{li2025promptminer,shen2024prompt}, framing the task as either white-box or black-box problem. White-box methods are 
gradient-based optimization approaches that require access to the model architecture and to the text and vision encoders. Black-box methods, in
contrast, assume access only to the target image, with no access to the text-to-image generative model. Our method (Figure~\ref{fig:placeholder}) is a
black-box approach: given only the target image, the objective is to recover the prompt that generated it. In this work, we focus on the black-box setting,
with particular emphasis on dense scenes.

\paragraph{Prompt Inversion}\label{subsec:prompt_inversion}
Prior prompt-inversion methods fall into three families, gradient-based approaches optimize prompt embeddings against a CLIP or denoising objective~\cite{mahajan2024prompting, wen2023hard} making recovered prompts drift off the natural-language manifold and are often unreadable. Autoregressive approaches decode prompts with a captioner or language model~\cite{li2022blip, clip_interrogator}, producing readable and interpretable text where claims are never checked against the image. RL based approaches formulate prompt recovery as a Markov Decision Process (MDP), training a policy over prompt tokens with image-similarity rewards~\cite{li2025promptminer} optimizing against the scoring model invites both evaluation circularity and overfitting to a single generator. In contrast, our method is bottom-up and training-free, rather than optimizing token sequences or fine-tuning any model, we are composing the prompt from the scene's content itself where detected instances, their attributes, and their spatial relations by simply using frozen, off-the-shelf VLMs. Further, we retain only those claims that pass an explicit per-claim verification test against the image. 

To capture the semantic context and its spatial relations, in this work
we break the image into multiple segments $S_1, S_2, \ldots, S_n$. Each
of these segments contributes toward the objective, where $\mathcal{C}(I)$
denotes the visual features and $\mathcal{C}(P)$ the textual features or
claims asserted by the prompt. We then define our objective as precision-
constrained recall maximization. In simple words, it means a prompt
maximizes the recoverable evidence to regenerate the image:
$\mathcal{P}(I) = \arg\max \mathcal{R}(I; P)$ such that
$\mathcal{C}(P) \subseteq \mathcal{C}(I)$.

To understand the composition and curation of the prompt, we divide the task in four stages (Algorithm~\ref{alg:pipeline}). a) Grounded Claim Proposal b) Evidence Gated Verification c) Geometric Fact Extraction, and d) Prompt Composition.

\paragraph{Grounded Claim Proposal} 

For creation of this framework, we start by curating the \textbf{proposal set}. To maximize the text extraction capability of captioning models or VLMs, we initialize the set of models with VLMs $g_{\phi}$. The input to the VLM is the image and the caption to extract the initial prompt. $c_{g} = g_{\phi}(I, \pi_{cap})$. Once we have obtained the caption, we need a query set $Q$. A query set is needed for the detector, as we use an open vocabulary detector. This open vocabulary detector cannot detect everything in the prompts, so it needs a recall ceiling for the whole pipeline. This object not present in the query set is never detected, never cropped, never described, and can never appear in the prompt. $Q$ is presented as the union of two sets of nouns. The first noun set is extracted from the global captions $N(c_{g})$, the other noun set $N(g_{\phi}(I, \pi_{inv}))$ comes from the inventory pass that extracts every distinct visible object, and this presents the formalization as in equation~\ref{eq:query_Set}:

\begin{equation}\label{eq:query_Set}
    Q = N(c_{g}) \cup N(g_{\phi}(I, \pi_{inv}))
\end{equation}

Once a complete query set has been abstracted, we present the query set $Q$ along with the image $I$ to the open vocabulary object detector $D_{\theta}$. This open vocabulary object detector presents us with the bounding boxes $b_{i}$ and the label $l_{i}$. More precisely $\{(b_{i}, l_{i}) \} = D_{\theta}(I, Q)$. Once the bounding boxes are extracted, this is fed to the crop operator $C_{p}$ along with the actual image $I$. This is extracted as $x_{i} = C_{p}(I, \rho(b_{i}))$. We do not directly use $b_{i}$ as the input to the crop operator, rather we map it back to the original coordinates using $\rho$, let us say we have a working copy of the image 512 px, thus the bounding box $b_{i}$ is projected as $\rho(b_i) = \left( x_0 \frac{W}{512}, y_0 \frac{H}{512}, x_1 \frac{W}{512}, y_1 \frac{H}{512} \right)$, where $W \times H$ is the original size of the image. The detection is robust to affine transforms as well as scale changes but the generation of description are not robust to these transformations since localization survives downsampling while fine-grained description does not, thus we reproject it back to the original size. The $p$ in $C_{p}$ denotes the relative context padding, this is important as zero context padding or a tight crop can make the context ambiguous. Once we have obtained $x_{i}$, we feed this to the VLM $g_{\phi}$ to obtain the slot schema description $d_{i} = g_{\phi}(x_{i}, \pi_{slot} | l_{i})$

\begin{table*}[t]
\centering
\caption{Image similarity across datasets, averaged over two seeds. Higher is better for DINO ($\uparrow$) and CLIP ($\uparrow$); lower is better for LPIPS ($\downarrow$). \textbf{Bold} marks the best result per column within each dataset. Our row is highlighted.}
\label{tab:main_comparison}
\small
\setlength{\tabcolsep}{13pt}
\renewcommand{\arraystretch}{1.12}
\begin{tabular}{ll ccc ccc}
\toprule
& & \multicolumn{3}{c}{\textbf{FLUX.1 dev}}
  & \multicolumn{3}{c}{\textbf{Stable Diffusion 3.5 Large}} \\
\cmidrule(lr){3-5}\cmidrule(lr){6-8}
\textbf{Dataset} & \textbf{Method}
  & DINO$\uparrow$ & LPIPS$\downarrow$ & CLIP$\uparrow$
  & DINO$\uparrow$ & LPIPS$\downarrow$ & CLIP$\uparrow$ \\
\midrule
\midrule
\multirow{8}{*}{\textit{MS COCO}}
 & BLIP           & 0.426 & 0.686 & 0.703 & 0.480 & 0.692 & 0.727 \\
 & CLIP-IG        & 0.520 & 0.679 & 0.716 & 0.505 & 0.710 & 0.727 \\
 & VLM-as-expert  & 0.578 & 0.660 & 0.754 & 0.597 & 0.679 & 0.763 \\
 & PH2P           & 0.033 & 0.744 & 0.410 & 0.024 & 0.761 & 0.427 \\
 & STEPS          & 0.359 & 0.691 & 0.677 & 0.419 & 0.706 & 0.736 \\
 & VGD            & 0.357 & 0.690 & 0.674 & 0.402 & 0.696 & 0.713 \\
 & PromptMiner    & 0.518 & 0.755 & 0.760 & 0.533 & 0.684 & 0.750 \\
 & \textsc{PROVE}  & \textbf{0.660} & \textbf{0.647} & \textbf{0.790} & \textbf{0.615} & \textbf{0.664} & \textbf{0.775} \\
\midrule
\multirow{8}{*}{\textit{Flickr30K}}
 & BLIP           & 0.382 & 0.689 & 0.670 & 0.401 & 0.690 & 0.672 \\
 & CLIP-IG        & 0.466 & 0.680 & 0.659 & 0.437 & 0.739 & 0.663 \\
 & VLM-as-expert  & 0.573 & 0.657 & 0.722 & 0.587 & 0.679 & 0.746 \\
 & PH2P           & 0.033 & 0.736 & 0.396 & 0.027 & 0.761 & 0.434 \\
 & STEPS          & 0.341 & 0.692 & 0.654 & 0.410 & 0.692 & 0.701 \\
 & VGD            & 0.378 & 0.690 & 0.659 & 0.416 & 0.689 & 0.698 \\
 & PromptMiner    & 0.518 & 0.684 & 0.730 & 0.490 & 0.693 & 0.724 \\
 & \textsc{PROVE}  & \textbf{0.610} & \textbf{0.635} & \textbf{0.763} & \textbf{0.593} & \textbf{0.664} & \textbf{0.765} \\
 \midrule
\multirow{8}{*}{\textit{Lexica}}
 & BLIP           & 0.535 & 0.615 & 0.755 & 0.480 & 0.649 & 0.744 \\
 & CLIP-IG        & 0.607 & 0.599 & 0.790 & 0.587 & 0.621 & 0.841 \\
 & VLM-as-expert  & 0.621 & 0.598 & 0.770 & 0.627 & 0.607 & 0.822 \\
 & PH2P           & 0.137 & 0.700 & 0.507 & 0.101 & 0.727 & 0.532 \\
 & STEPS          & 0.540 & 0.615 & 0.784 & 0.515 & 0.628 & 0.809 \\
 & VGD            & 0.468 & 0.627 & 0.761 & 0.420 & 0.646 & 0.762 \\
 & PromptMiner    & \textbf{0.742} & \textbf{0.560} & \textbf{0.890} & 0.596 & 0.613 & 0.826 \\
 & \textsc{PROVE}  & 0.741 & \textbf{0.560} & 0.853 & \textbf{0.611} & \textbf{0.612} & \textbf{0.837} \\
\bottomrule
\end{tabular}
\end{table*}

\paragraph{Evidence Gated Verification}

This step is used as a verification for the claims that are construed by the proposer. After we obtain the slot schema description $d_{i}$ that are conditioned on the labels $l_{i}$. These claims are verified using a gated mechanism where the gate is an anchored evidence gate used using a function, in our case we have used CLIP as the function. In this step we first have the segmenter, followed by the evidence function with the gate, and finally the kept-set.

The segmenter first divides the segmentation into further atomic claims. This is needed as the text $d_{i}$ comes as a free flowing textual paragraph, which are difficult to verify. A big paragraph description can make verification difficult as singular score for the whole paragraph as it would let one false clause hide among true ones, or one false clause sink the paragraph. Thus we divide the claims using a segmenter. The segmenter here is not a learned model but a deterministic function based on common splits. S is a deterministic rule-based segmenter that approximates syntactic phrase boundaries using closed-class vocabulary, a compact visual-adjective lexicon, and suffix morphology. This is given by the $S(d_{i}) = (c_{1}, ....c_{m})$. Each of these claims $c_{i}$ is investigated against its own regions using a label-relative evidence function. This is given in equation~\ref{eq:evidence_function} where $f$ is the CLIP model and $A$ is the anchoring operator. The role of the anchoring operator determines whether the attribute fragments competes fairly against the noun dominated labels.
\begin{equation} \label{eq:evidence_function}
\begin{split}
    E(c;x_{i},l_{i}) &= sim(f_{I}(x_{i}), f_{T}(A(c_{i}, l_{i}))) \\
    &\quad - sim(f_{I}(x_{i}), f_{T}(l_{i}))
\end{split}
\end{equation}

The kept-set $K_{i}$, are the claims whose evidence clears the threshold. The threshold is given by the threshold gate $\delta$, given by $K_{i} = \{ c \in S(d_{i} : E(c) \ge -\delta)$. 
The threshold presented here is negative as the baseline is the label, thus the demanding claims sits above the baseline $E \ge 0$ makes it too strict. This is as we use the CLIP, the CLIP similarity is dominated by the object or category noun. The score decreases as we tend to increase the length and become more specific. The specificity signal in the prompts decrease the score as it dilutes the object or category signal~\cite{zhang2024longclip, Urbanek_2024_CVPR}. This can be simply thought of as $\mathbf{sim}(claim) \ge \mathbf{sim}(l_{i}) - \delta $.

\paragraph{Geometric Fact Extraction}
This part is exempted from evidenced gated verification, as this is a  grounded deterministic mechanism. These facts are computed directly using the $\{(b_{i}, l_{i})\}$, using the deterministic assumptions. In this part, ground truth  is the detector output itself. First, let us say that the count $cnt$ is given by $\mathbf{cnt}_{l} = \bigl|\{\, o : l_{o} = l \,\}\bigr|$, thus for each label, the tally of the detections is determined by the NMS or spatial dedup. For labels with $n_{l} \ge 2$ the prose is directly used in the final prompt, which is determined at the composition level. The support relations $\mathbf{rel_d}(o, u)$ are then computed in the describer, post-verification, over the scene parts. More commonly, this is specifically a surface-band rule over the bounding boxes. Finally, we compute the inter instance spatial relations $\mathbf{rel_s}(o, o')$ that bypasses the evidence gate by the design. The spatial relations are presented at a level where it is computed by sorting the object centroid, adjacent pairs in order to get the relations by normalized gap, and finally the holding of the object i.e, any instance whose assigned parts include a held-category label (eg: man on the \textit{left} is holding the controller).  Since the evidence gate designed on top of the CLIP, it is unable to capture the global fine-grained details, contrastive image-text models behave similar to the bag-of-words on counting and the spatial order~\cite{yuksekgonul2023when, basu-etal-2024-distilling}.

\subsection{Prompt Composition}
This is the final stage for the construction of the prompt that we feed to text-to-image model. The composition stage is divided into two different parts: a) verified hierarchy, and b) deterministic serialisation.


The verified hierarchy is a pair $H(I) = \langle \textit{global},\, \textit{instances}\rangle$. The enriched global text combines the caption
$c_g$, verified scene attributes, counts, and injected relations where each instance carries its label, position tag, box, and kept claim-set $K_k$. This structure is auditable by construction where every claim is grounded and carries its evidence score.

Text-to-image generators, however, are trained on flat text and cannot ingest hierarchy, so $H(I)$ must be linearized. This is the role of the deterministic composition operator $\mu$. It emits the global text first, then per-instance positional clauses, and finally a background clause for unowned scene objects. A repair pass removes dangling fragments and duplicate claims and rejoins the result into fluent sentences.

\section{Experiments}

We design experiments to measure image recovery for original real images and AI-generated images. For the original real images domain, we use the MS-COCO~\cite{lin2014microsoft} and Flickr30K~\cite{plummer2015flickr30k} datasets. MS-COCO and Flickr30K provide real photographs paired with human-written captions, and stress prompt recovery on \emph{dense, everyday scenes} that contain multiple interacting objects, counts, and spatial relations, exactly the regime our per-claim evidence pipeline targets. For MS-COCO and Flickr30K a prompt is generated directly from original dataset images.

For the generative-art domain, we use the Lexica~\cite{lexica} dataset. Lexica consists of art-style images  paired with the engineered prompts that produced them, and stresses recovery of the \emph{stylistic and aesthetic modifiers}
characteristic of prompt marketplaces. 

For Lexica, following prior work, we use the provided ground-truth prompt to generate an image, which in turn used to generate the inverted prompt. We use two state-of-the-art open-weight text-to-image models both to render recovered prompts back into images and, where noted above, to construct the target images: FLUX.1-dev~\cite{labs2025flux1kontextflowmatching} and Stable Diffusion~3.5 Large~\cite{esser2024scaling}. Each method is evaluated independently using both generators. Crucially, the generator is used only to \emph{render} images for evaluation and is applied identically to all methods.





We compare against seven prompt-inversion methods drawn from the three
established families. \textbf{Optimization-based:}
PH2P~\cite{mahajan2024prompting}, STEPS~\cite{Qiu2025STEPSSP}, and
VGD~\cite{kim2025visually}, which recover prompts by gradient-based or
guided search against a similarity objective. \textbf{Captioning-based:}
BLIP~\cite{li2022blip}, CLIP-Interrogator (CLIP-IG)~\cite{clip_interrogator}, and a strong VLM-as-expert baseline that prompts a frozen vision-language model to describe the image directly. \textbf{RL-based:} PromptMiner~\cite{li2025promptminer}, which learns a prompt-recovery policy via reinforcement learning with fuzz-based modifier optimization. All methods take a single source image as input and return text. Since prompt inversion is ultimately judged by whether the recovered prompt reproduces the target, we report metrics comparing the resulting image to the target image: \textbf{DINO}~($\uparrow$), the cosine similarity of self-supervised DINO features; \textbf{LPIPS}~($\downarrow$), a learned perceptual distance, and \textbf{CLIP}~($\uparrow$), the CLIP-space cosine similarity between the recovered prompt and the target image (text-image alignment). Reporting a semantic image measure (DINO), a perceptual image measure (LPIPS), and a text-image alignment measure (CLIP) guards against any single metric being gamed by an off-manifold or degenerate prompt. All generations are done twice using two seeds, and average metrics of the two generations are reported. As shown in Table~\ref{tab:main_comparison}, PROVE achieves the strongest overall image recovery across all three datasets, with qualitative reconstructions in Figure~\ref{fig:qual} confirming that it recovers object counts and spatial layout more faithfully than the baselines.



\paragraph{Results.} Table~\ref{tab:main_comparison} reports image-similarity metrics across 100 randomly selected images from each of the three datasets, across two text-to-image generators (\emph{FLUX.1-dev} and \emph{Stable Diffusion~3.5~Large}). PROVE is compared against seven state-of-the-art white-box and black-box methods. \textsc{PROVE} provides new state-of-the-art results. While Table~\ref{tab:main_comparison} presented aggregate quantitative results for 600 images across the three datasets and two generators, Figure~\ref{fig:qual} presents sample qualitative examples from the three datasets. Additional examples are provided in the supplementary material.
 
 
 
 
\section{Robustness and Defenses}
Because prompt inversion can be misused, we evaluate \textsc{PROVE} under defenses an image owner might deploy: three post-hoc perturbations from prior
work~\cite{shen2024prompt} such as \textbf{Noise Injection}, \textbf{Puzzle}, and \textbf{Watermarking}, as well as a stronger white-box defense, \textbf{Adversarial CLIP}, which perturbs the image against the CLIP encoder while leaving it
perceptually intact. Results are reported in Table~\ref{tab:defense_flux}.

\textbf{Noise Injection} adds zero-mean Gaussian noise to the target before inversion, degrading low-level statistics without altering scene content, and
thus probes whether inversion depends on precise pixel appearance, a method whose information flows through semantic channels (captioning, detection, region description) should be unaffected, whereas one keyed to fine texture should degrade. \textbf{Puzzle} divides the image into a $4\times4$ grid and randomly offsets each patch, disrupting global spatial coherence while leaving local content intact. It perturbs scene \emph{composition} rather than appearance and so tests our geometric-fact channel, where object identity survives within patches, so counts persist, while support and relation predicates, computed from box coordinates, are most exposed to the displacement. \textbf{Watermarking} puts semi-transparent text across the image. Unlike the previous two, it \emph{adds} content rather than removing it, probing whether verification suppresses inserted content lacking a grounded visual reference.

Following the white-box formulation of~\cite{shen2024prompt}, \textbf{Adversarial CLIP} applies iterative FGSM to the CLIP image encoder, minimizing cosine
similarity between perturbed and clean embeddings within an $L_\infty$ ball ($\epsilon=8/255$, 100 steps) -- the standard bound in the adversarial-robustness literature, at which the perturbation is imperceptible in natural images. Prior inversion methods score, search, and evaluate through the same encoder, so the perturbation corrupts their objective end to end. Our decomposition into a CLIP-free proposal stage and a CLIP-guided verification stage confines the effect to the verification gate alone: the VLM and detector channels are unaffected, and unverified claims fall back to their object labels. The prompt therefore degrades toward reduced specificity rather than toward content uncorrelated with the image.

\begin{table}[t]
\centering
\caption{Image similarity across datasets for \textbf{FLUX.1 dev}, averaged over two seeds. Results are reported against each defense to show the robustness of the attack. Higher is better for DINO ($\uparrow$), CLIP ($\uparrow$), and SBERT ($\uparrow$); lower is better for LPIPS ($\downarrow$).}
\label{tab:defense_flux}
\small
\setlength{\tabcolsep}{6pt}
\renewcommand{\arraystretch}{1.16}
\begin{tabular}{ll ccc}
\toprule
\textbf{Dataset} & \textbf{Method}
  & \textbf{DINO}$\uparrow$ & \textbf{LPIPS}$\downarrow$ & \textbf{CLIP}$\uparrow$ \\
\midrule
\midrule
\multirow{5}{*}{\textit{MS COCO}}
 & Clean          & 0.660 & 0.647 & 0.790 \\
 & Noise          & 0.632 & 0.658 & 0.779 \\
 & Puzzle         & 0.635 & 0.656 & 0.783 \\
 & Watermark      & 0.566 & 0.666 & 0.750 \\
 & Adv. CLIP      & 0.654 & 0.655 & 0.787 \\
\midrule
\multirow{5}{*}{\textit{Flickr30K}}
 & Clean          & 0.610 & 0.635 & 0.763 \\
 & Noise          & 0.583 & 0.656 & 0.741 \\
 & Puzzle         & 0.619 & 0.651 & 0.754 \\
 & Watermark      & 0.551 & 0.660 & 0.715 \\
 & Adv. CLIP      & 0.604 & 0.652 & 0.750 \\
\midrule
\multirow{5}{*}{\textit{Lexica}}
 & Clean          & 0.741 & 0.560 & 0.853 \\
 & Noise          & 0.708 & 0.570 & 0.823 \\
 & Puzzle         & 0.712 & 0.568 & 0.837 \\
 & Watermark      & 0.657 & 0.590 & 0.776 \\
 & Adv. CLIP      & 0.720 & 0.570 & 0.843 \\
\bottomrule
\end{tabular}
\end{table}

\section{Conclusion}
We propose \textsc{PROVE}, a training-free, optimization-free prompt inversion (prompt stealing) method that recovers the prompt used to generate a target
image. \textsc{PROVE} proceeds in four stages. First, a frozen VLM and open-vocabulary detection propose fine-grained atomic claims about the image.
Second, each claim is retained only if it satisfies a label-anchored CLIP margin on its corresponding region. Third, counts and spatial relations are derived deterministically from the detections rather than asserted by the VLM. Finally, the verified claims are serialized into a single fluent prompt, with no generator in the loop. \textsc{PROVE} recovers more descriptive and accurate prompts than the considered baselines. These gains come from verification and deterministic derivation, without any training, optimization, or access to the target generator. While this constitutes a stronger prompt-stealing attack, with direct implications for copyright and the reproduction of content, our pipeline verifies visual content while, by construction, refusing artist and platform attribution, separating faithful description from unauthorized reproduction of an artist's identity. PROVE reports visual evidence without asserting authorship. We hope this separation of description from identity offers a useful reference point for studying copyright concerns.
\clearpage
\bibliography{aaai2027}

\newpage
\clearpage
\section{\LARGE Supplementary Materials}

This supplementary material provides additional analyses and results that complement the main paper. We present extended qualitative examples across datasets, evaluate how well the attack generalizes to different generators, examine prompt alignment, assess attack robustness on SD 3.5, and report implementation details.

\section{Additional Qualitative Examples}
Figure~\ref{fig:qual-supp} shows additional qualitative results on images sampled from MS-COCO, Flickr30K, and Lexica, spanning both natural and generated images. Figures~\ref{fig:qual_coco}, \ref{fig:qual_flickr}, and \ref{fig:qual_lexica} compare the prompts produced by PROVE against those from all seven competing methods, alongside the images generated from each prompt.

\section{Attack Robustness on SD 3.5}
The main paper presents attack-robustness results on FLUX.1-dev. Table~\ref{tab:defense_sd35} reports the same experiment on SD3.5-Large. The trend is consistent, \textsc{PROVE} shows a clear improvement over the baselines, despite PromptMiner being directly optimized for CLIP score.

\section{Prompt Alignment}

Lexica provides an initial starting prompt (ground truth), which lets us measure the text alignment of the recovered prompt against it. We quantify this alignment using Sentence-BERT (SBERT) similarity. \textbf{SBERT}~($\uparrow$) computes the cosine similarity between the recovered prompt and the ground-truth prompt. We report SBERT on the Lexica dataset against all baselines, and additionally show the clean responses under other defenses. Table~\ref{tab:sbert_lexica} presents the results of this experiment. \textsc{PROVE} attains the best SBERT on Lexica ($0.470\,/\,0.493$ for the \emph{dense}\,/\,\emph{simple} splits, versus $\text{0.466}\,/\,\text{0.474}$). Consistent with 
image-text dissociation, a method that does not optimize caption-form similarity can trail on SBERT against the target captions while remaining more faithful to image content. 

Additionally, we measure textual alignment using SBERT similarity under different defense mechanisms. The resulting performance is reported in Table~\ref{tab:sbert_lexica_defense}.

\begin{table}[tp]
\centering
\caption{Image similarity across datasets for \textbf{SD 3.5 Large}, averaged over two seeds. Results are reported against each defense to show the robustness of the attack. Higher is better for DINO ($\uparrow$), CLIP ($\uparrow$), and SBERT ($\uparrow$); lower is better for LPIPS ($\downarrow$).}
\label{tab:defense_sd35}
\small
\setlength{\tabcolsep}{6pt}
\renewcommand{\arraystretch}{1.16}
\begin{tabular}{ll ccc}
\toprule
\textbf{Dataset} & \textbf{Method}
  & \textbf{DINO}$\uparrow$ & \textbf{LPIPS}$\downarrow$ & \textbf{CLIP}$\uparrow$ \\
\midrule
\midrule
\multirow{5}{*}{\textit{MS COCO}}
 & Clean          & 0.615 & 0.664 & 0.775  \\
 & Noise           & 0.603 & 0.664 & 0.765  \\
 & Puzzle          & 0.610 & 0.667 & 0.771  \\
 & Watermark       & 0.540 & 0.680 & 0.740 \\
 & Adv. CLIP       & 0.613 & 0.672 & 0.771 \\
\midrule
\multirow{5}{*}{Flickr30K}
 & Clean          & 0.593 & 0.664 & 0.765  \\
 & Noise           & 0.574 & 0.667 & 0.745  \\
 & Puzzle          & 0.597 & 0.661 & 0.759  \\
 & Watermark       & 0.540 & 0.673 & 0.727 \\
 & Adv. CLIP      & 0.578 & 0.663 & 0.757 \\
 \midrule
\multirow{5}{*}{Lexica}
 & Clean           & 0.611 & 0.612 & 0.837  \\
 & Noise           & 0.608 & 0.630 & 0.817  \\
 & Puzzle         & 0.604 & 0.653 & 0.827   \\
 & Watermark       & 0.555 & 0.624 & 0.787  \\
 & Adv. CLIP       & 0.600 & 0.644 & 0.833  \\
\bottomrule
\end{tabular}
\end{table}

\begin{table}[tp]
\centering
\caption{Textual alignment (SBERT$\uparrow$) on the Lexica dataset, averaged over two seeds. \textbf{Bold} marks the best result per column.}
\label{tab:sbert_lexica}
\small
\setlength{\tabcolsep}{8pt}
\renewcommand{\arraystretch}{1.12}
\begin{tabular}{l cc}
\toprule
& \textbf{FLUX.1 dev} & \textbf{SD 3.5 Large} \\
\textbf{Method} & SBERT$\uparrow$ & SBERT$\uparrow$ \\
\midrule
\midrule
BLIP           & 0.288 & 0.306 \\
CLIP-IG        & 0.328 & 0.351 \\
VLM-as-expert  & 0.466 & 0.474 \\
PH2P           & 0.154 & 0.154 \\
STEPS          & 0.372 & 0.392 \\
VGD            & 0.405 & 0.411 \\
PromptMiner    & 0.467 & 0.481 \\
\textsc{PROVE}  & \textbf{0.470} & \textbf{0.493} \\
\bottomrule
\end{tabular}
\end{table}

\begin{table}[tp]
\centering
\caption{Textual alignment (SBERT$\uparrow$) on the Lexica dataset, averaged over two seeds. This is computed against the various defense mechanisms used to test the robustness of the attack.}
\label{tab:sbert_lexica_defense}
\small
\setlength{\tabcolsep}{8pt}
\renewcommand{\arraystretch}{1.12}
\begin{tabular}{l cc}
\toprule
& \textbf{FLUX.1 dev} & \textbf{SD 3.5 Large} \\
\textbf{Method} & SBERT$\uparrow$ & SBERT$\uparrow$ \\
\midrule
\midrule
Clean          & 0.470 & 0.493 \\
Noise          & 0.447 & 0.457 \\
Puzzle         & 0.453 & 0.482 \\
Watermark      & 0.446 & 0.462 \\
\textsc{Adv. CLIP}  & {0.454} & {0.480} \\
\bottomrule
\end{tabular}
\end{table}

\begin{table*}[tp]
\centering
\caption{Image-similarity metrics across datasets, averaged over two seeds. Higher is better for DINO~($\uparrow$) and CLIP~($\uparrow$); lower is better
for LPIPS~($\downarrow$). \textbf{Bold} marks the best result per column within each dataset. In this cross-test, the target image and the image re-rendered from the recovered prompt are produced by \emph{different} generators.}
\label{tab:cross-test}
\small
\setlength{\tabcolsep}{13pt}
\renewcommand{\arraystretch}{1.12}
\begin{tabular}{ll ccc ccc}
\toprule
& & \multicolumn{3}{c}{\textbf{FLUX.1 dev}}
  & \multicolumn{3}{c}{\textbf{Stable Diffusion 3.5 Large}} \\
\cmidrule(lr){3-5}\cmidrule(lr){6-8}
\textbf{Dataset} & \textbf{Method}
  & DINO$\uparrow$ & LPIPS$\downarrow$ & CLIP$\uparrow$
  & DINO$\uparrow$ & LPIPS$\downarrow$ & CLIP$\uparrow$ \\
\midrule
\midrule
\multirow{2}{*}{\textit{Lexica}}
 & PromptMiner    & {0.593} & {0.610} & \textbf{0.850} & 0.615 & 0.618 & 0.817 \\
 & \textsc{PROVE}  & \textbf{0.645} & \textbf{0.590} & 0.839 & \textbf{0.641} & \textbf{0.604} & \textbf{0.827} \\
\bottomrule
\end{tabular}
\end{table*}

\section{Attack Generalization to Different Generators}
MS-COCO and Flickr30K start from real images, whereas Lexica images are generated from a given prompt. Since a generated image depends on the generator
that produced it, we test our attack in the setting where the generator used for the original image differs from the one used during the attack. In addition, we cross-test PromptMiner and \textsc{PROVE} in Table~\ref{tab:cross-test}. Using prompts sourced from Lexica, we generate the target images with SD 3.5 Large and FLUX.1-dev, and recover prompts that are then re-rendered with the opposite generator i.e, FLUX.1-dev for the SD 3.5 Large
targets and SD 3.5 Large for the FLUX.1-dev targets. Across both settings, \textsc{PROVE} achieves stronger performance.

\newcommand{\qualgridsupp}{%
  \begin{minipage}{\figwidth}
    \begin{tabular}{@{}c*{9}{c}@{}}
      \colhead
      \gridrow{}{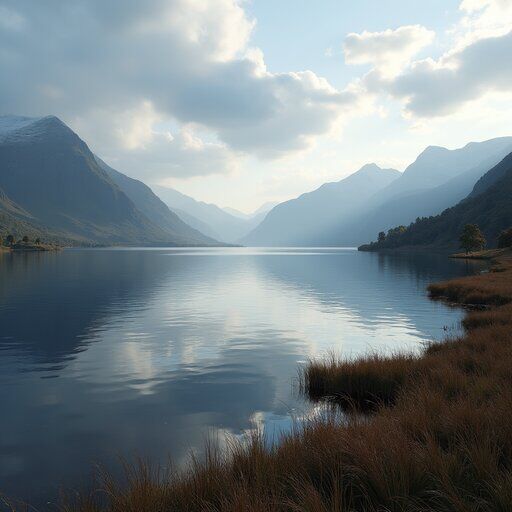}{jpg}
      \gridrow{MS-COCO}{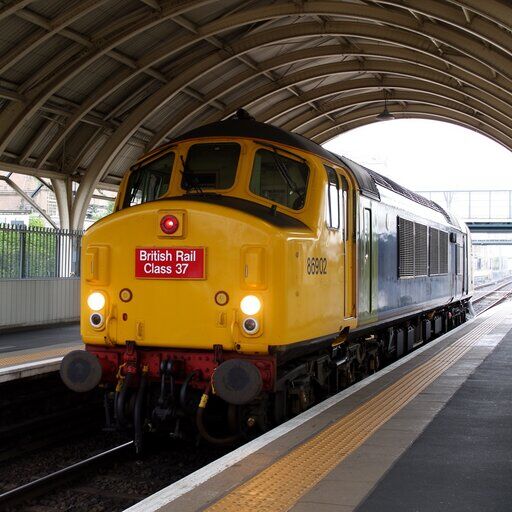}{jpg}
      \gridrow{}{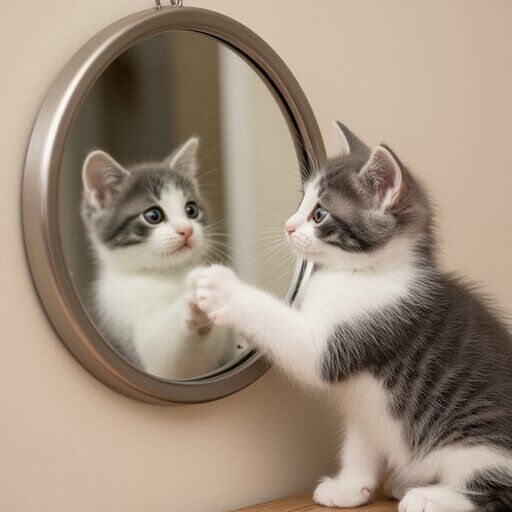}{jpg}
      \noalign{\vskip 7pt}
      \gridrow{}{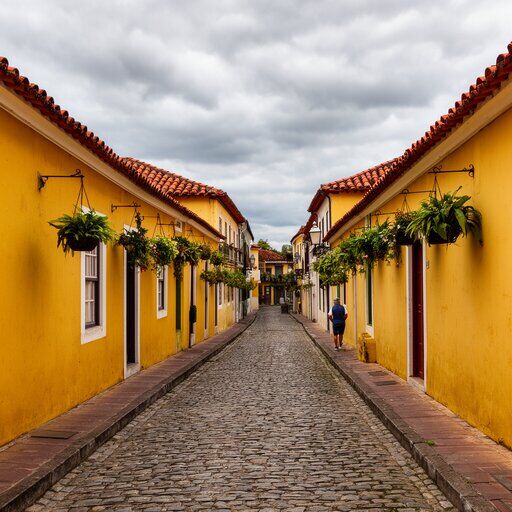}{jpg}
      \gridrow{Flickr}{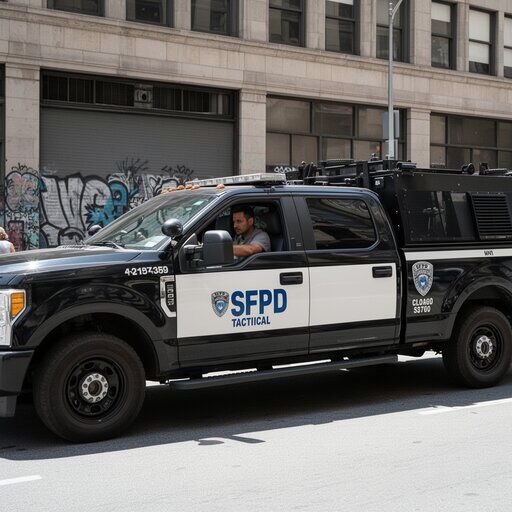}{jpg}
      \gridrow{}{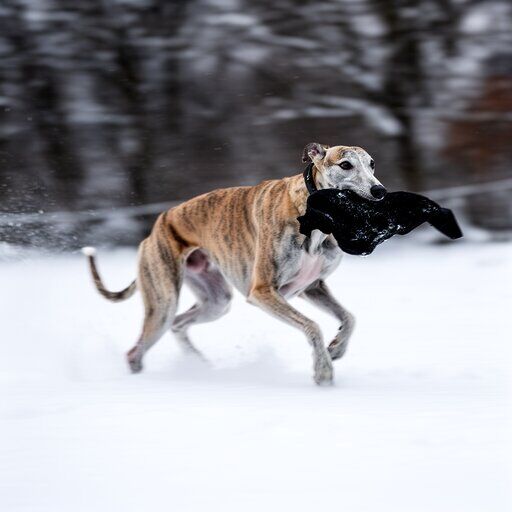}{jpg}
      \noalign{\vskip 7pt}
      \gridrow{}{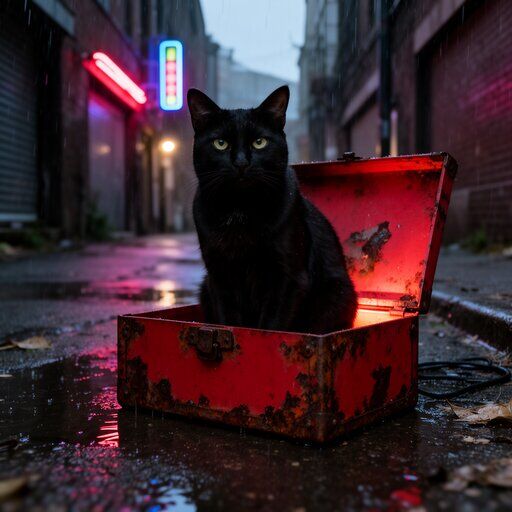}{jpg}
      \gridrow{Lexica}{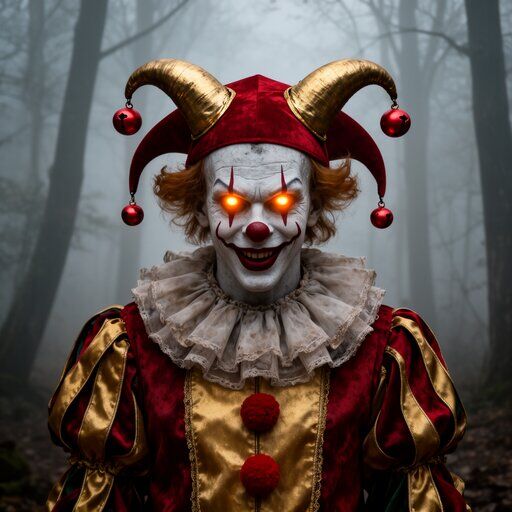}{jpg}
      \gridrow{}{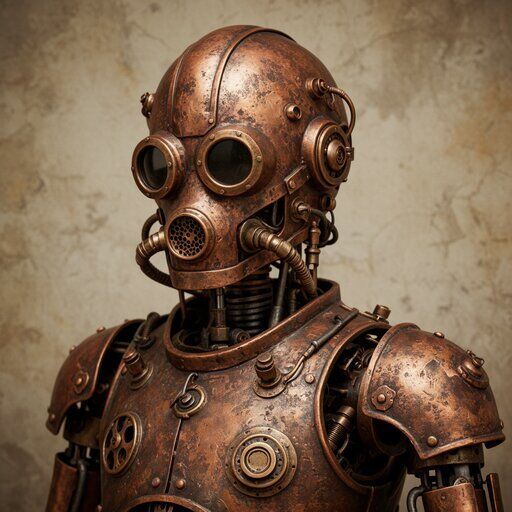}{jpg}
    \end{tabular}
  \end{minipage}%
}

\begin{figure*}[p]
  \centering
  \resizebox{\textwidth}{!}{\qualgridsupp}
  \caption{Additional qualitative comparisons on MS-COCO, Flickr30K, and Lexica.}
  \label{fig:qual-supp}
\end{figure*}

\begin{figure*}[t]
\centering
\begin{minipage}[c]{0.17\textwidth}
  \centering
  \includegraphics[width=\linewidth]{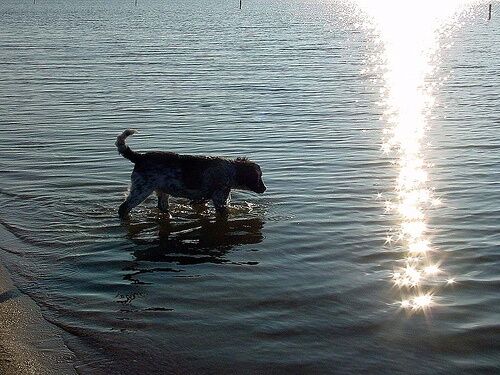}\\[4pt]
  \textbf{Target}
\end{minipage}\hfill
\begin{minipage}[c]{0.80\textwidth}
\renewcommand{\arraystretch}{1.3}
\setlength{\tabcolsep}{6pt}
\begin{tabular}{@{}>{\centering\arraybackslash}m{0.19\linewidth} m{0.75\linewidth}@{}}
\toprule
\textbf{Generated} & \textbf{Recovered prompt} \\
\midrule
\includegraphics[width=\linewidth]{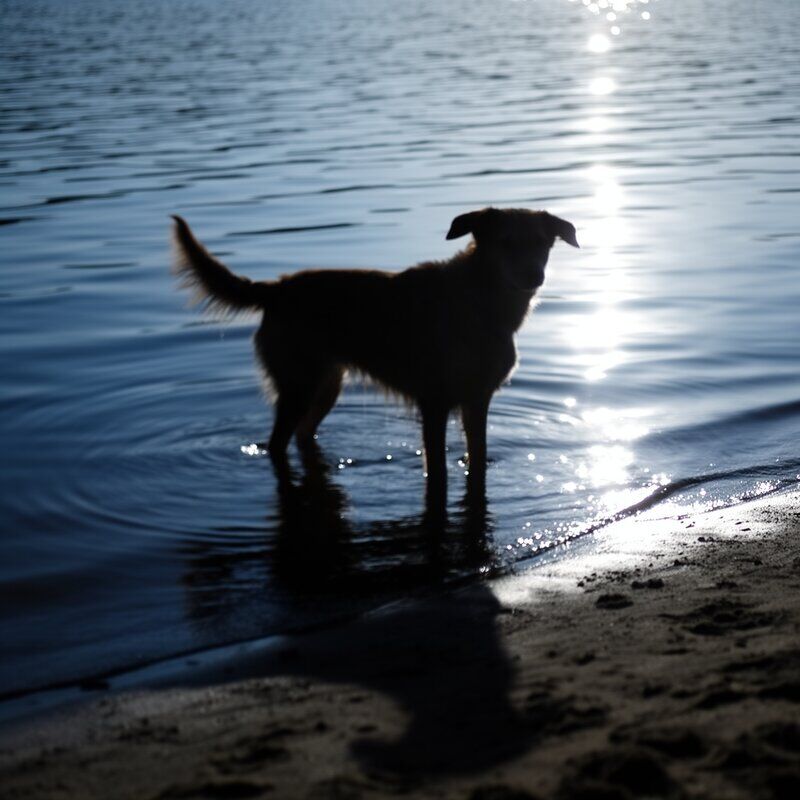} &
\textbf{Ours.} A dog stands in shallow water near a sandy shore, its silhouette highlighted by sunlight reflecting off the ripples of the lake. The scene is tranquil, with the dog's reflection visible on the wet sand. Bright, late afternoon, outdoor, eye-level, cool tones, blurred, water reflections. Standing in water, tail up, fur glistening, dark eyes, ears flopped down, legs submerged, water ripples around feet. Wet sand, dark water, ripples, reflections, shoreline. Distorted shapes, unclear details, smooth surface, no text, abstract form, no specific shape, no condition details. \\
\midrule
\includegraphics[width=\linewidth]{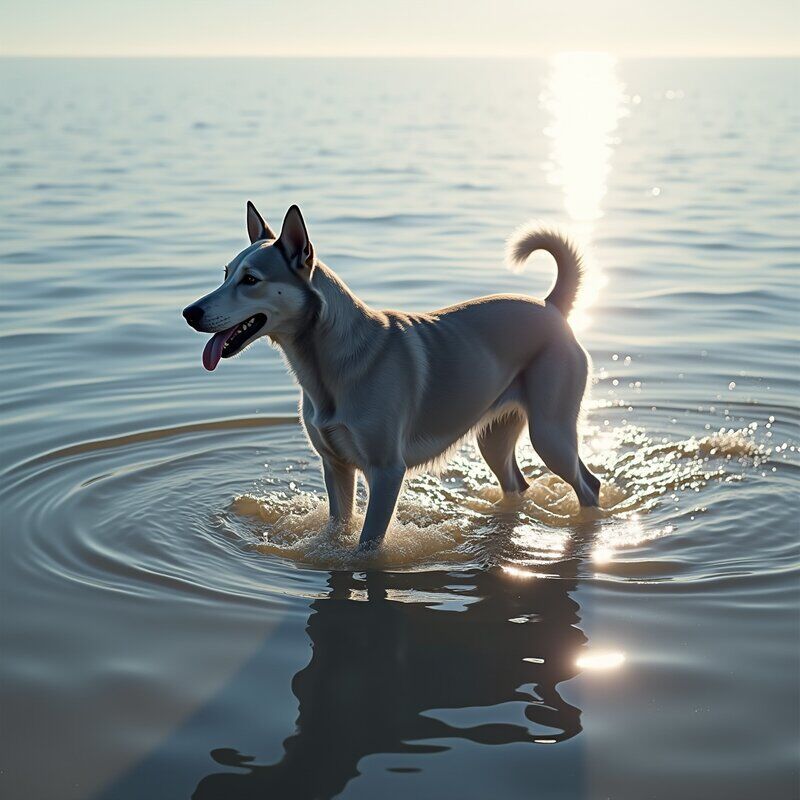} &
\textbf{PromptMiner.} a grey dog wading in calm water near the shore, reflecting in rippled water; sun reflection, body of calm water; clear day, isometric \\
\midrule
\includegraphics[width=\linewidth]{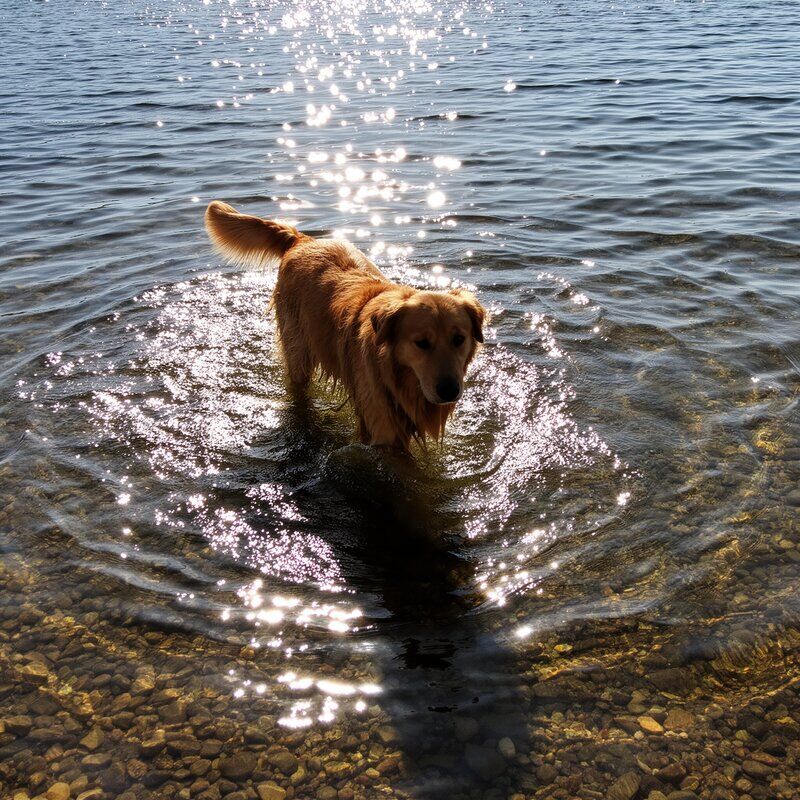} &
\textbf{VLM-as-expert.} A dog wading in shallow water with sunlight reflecting on the surface, creating sparkling patterns \\
\midrule
\includegraphics[width=\linewidth]{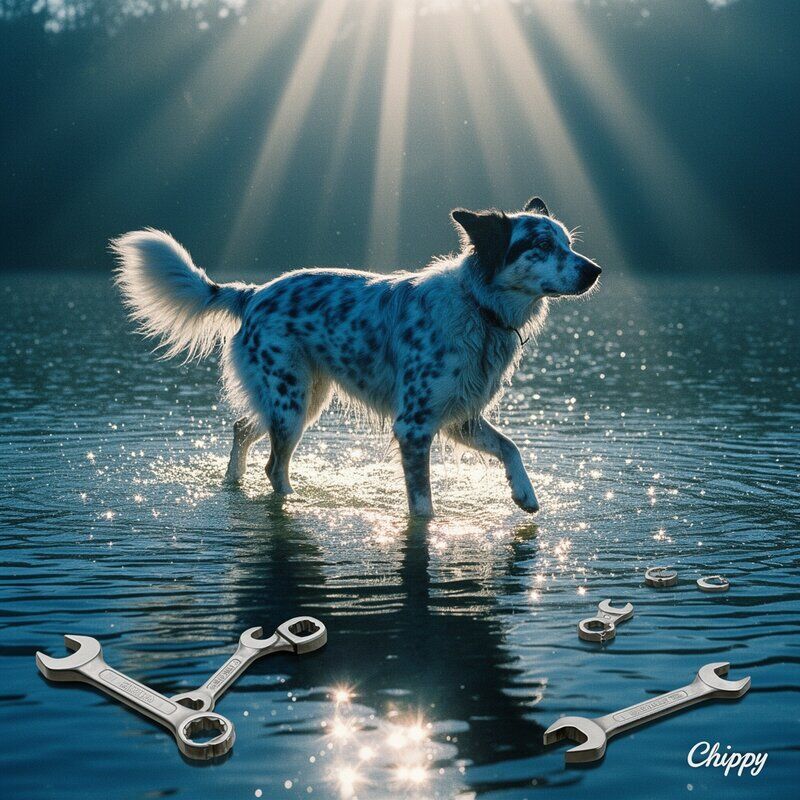} &
\textbf{CLIP-Interrogator.} a dog standing in the water, glittering sun rays, ebay photo, by Chippy, 1998 photo, reflections and refractions, backlighted, forced perspective. \\
\bottomrule
\end{tabular}
\end{minipage}
\caption{Qualitative comparison on Flickr30K. We compare \emph{PROVE} against its three closest baselines, showing both the generated image and the recovered prompt. On these natural, caption-style scenes, \emph{PROVE} recovers
finer-grained detailed prompts than the baselines, yielding reconstructions closer to the target.}
\label{fig:qual_flickr}
\end{figure*}

\begin{figure*}[t]
\centering
\begin{minipage}[c]{0.17\textwidth}
  \centering
  \includegraphics[width=\linewidth]{fig/ms-coco/target.jpg}\\[4pt]
  \textbf{Target}
\end{minipage}\hfill
\begin{minipage}[c]{0.80\textwidth}
\renewcommand{\arraystretch}{1.3}
\setlength{\tabcolsep}{6pt}
\begin{tabular}{@{}>{\centering\arraybackslash}m{0.19\linewidth} m{0.75\linewidth}@{}}
\toprule
\textbf{Generated} & \textbf{Recovered prompt} \\
\midrule
\includegraphics[width=\linewidth]{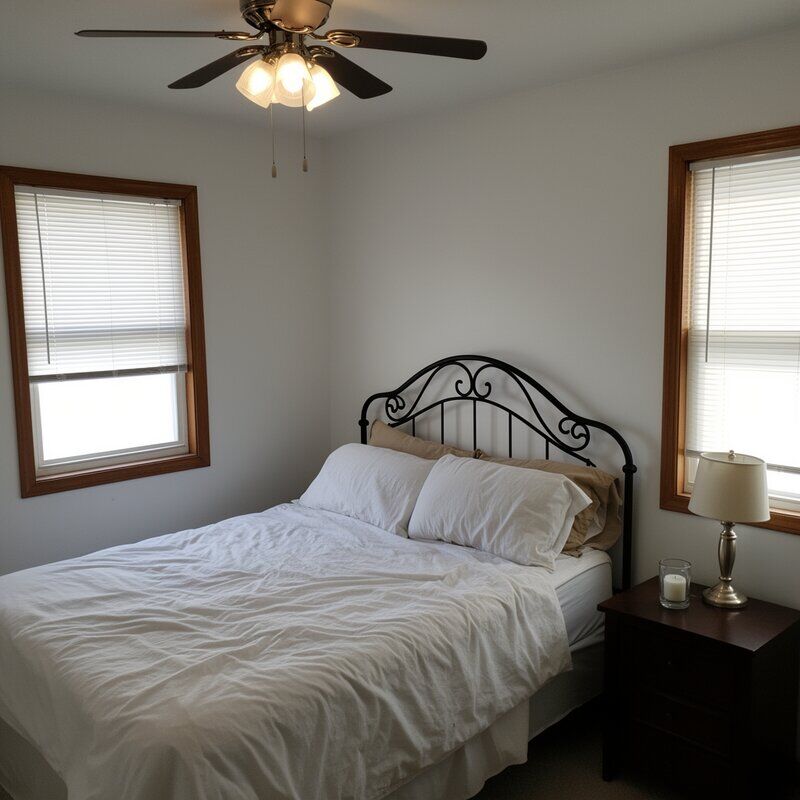} &
\textbf{Ours.} A bed with white linens and a metal headboard sits between two windows with closed blinds in a simple bedroom. A nightstand with a lamp and a candle is to the right of the bed. The room has a ceiling fan with lights. Bright, daytime, indoor, eye-level, neutral, plain. It contains three pillows, two fans. Black wrought iron headboard with curved scrollwork. Dark brown wooden nightstand, small glass candle holder, window with white frame, light source from window, shadowed corner, minimalistic. White pillow, soft, slightly wrinkled, on bed, unmade, indoor setting. Wooden frame, window blinds, nightstand lamp. Beige pillow with white pillowcase, slightly wrinkled, soft-looking, on bed with dark metal headboard. Clear glass, smooth surface, transparent, round, empty. Wooden-framed window with horizontal blinds, partially open, allowing natural light into room, white walls visible. \\
\midrule
\includegraphics[width=\linewidth]{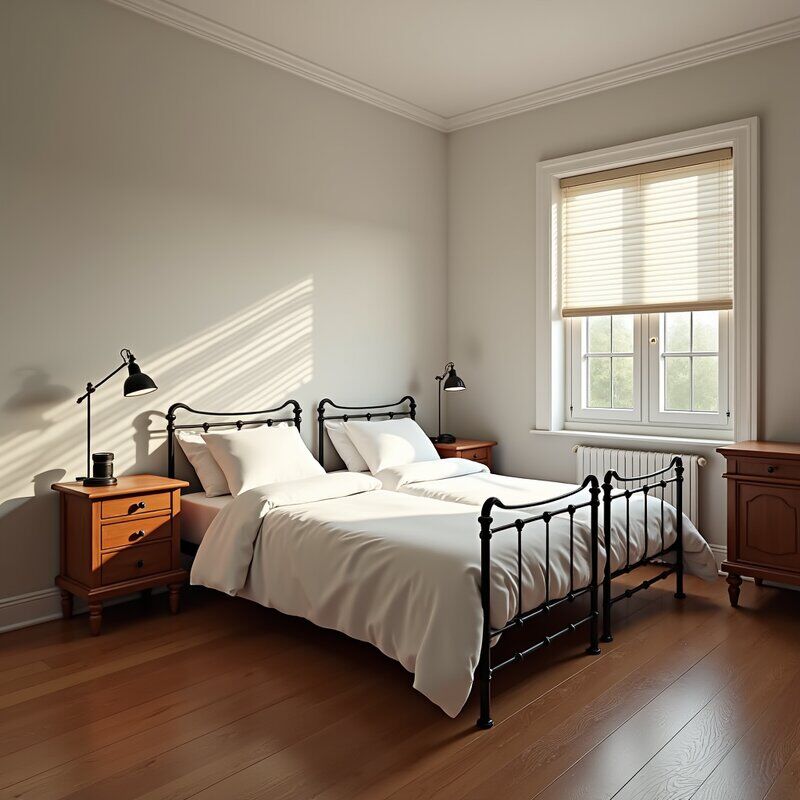} &
\textbf{PromptMiner.} bedroom, cosy bedroom, traditional wooden furniture, twin metal bed, white sheets, soft lighting, window with blinds, hardwood floor, low angle \\
\midrule
\includegraphics[width=\linewidth]{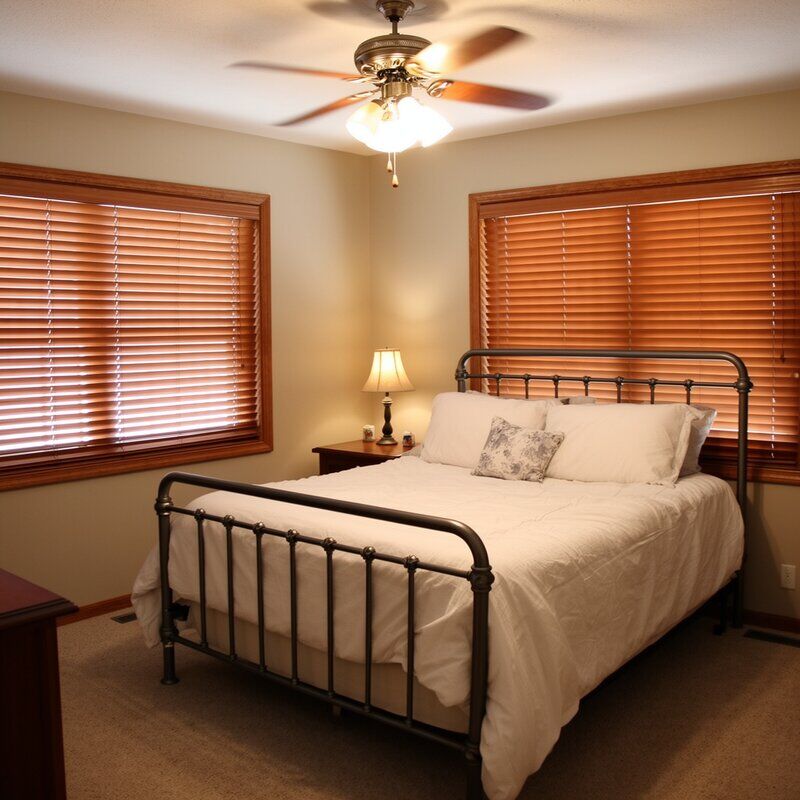} &
\textbf{VLM-as-expert.} A cozy bedroom with a metal bed frame, white bedding, wooden blinds, a nightstand with a lamp, and a ceiling fan with lights. \\
\midrule
\includegraphics[width=\linewidth]{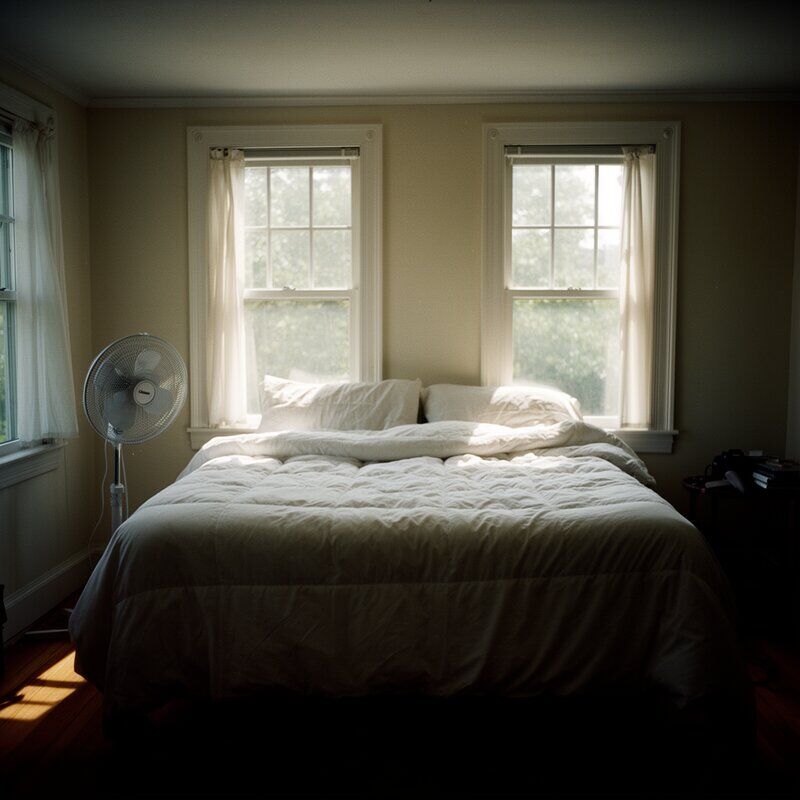} &
\textbf{CLIP-Interrogator.} a bed with a white comforter and a fan in a room, sunlight through windows, photo taken with canon 5d, by Gregory Gillespie, story board format.\\
\bottomrule
\end{tabular}
\end{minipage}
\caption{Qualitative comparison on MS-COCO. We compare \emph{PROVE} against its three closest baselines, showing both the generated image and the recovered
prompt. \emph{PROVE} recovers finer-grained, detailed prompts, which in turn yield reconstructions closer to the target.}
\label{fig:qual_coco}
\end{figure*}

\section{Implementation Details}

\paragraph{Models.}
All components are frozen, off-the-shelf models; \textsc{PROVE} requires no
training or gradient optimization. We use Qwen2.5-VL-7B-Instruct as the
vision-language model (VLM) $g_\phi$ for global captioning, scene-attribute
extraction, object inventory, and per-region slot-schema description.
Open-vocabulary detection $D_\theta$ uses OWLv2 (ViT-B/16, ensemble), and the
evidence gate $E$ scores claims with CLIP ViT-L/14. Region proposals are
deduplicated with non-maximum suppression at an IoU threshold of $0.5$.

\paragraph{Pipeline configuration.}
Detection runs on a $512\times512$ working copy, while all region crops are
taken at native resolution to preserve fine detail; crops are padded by a factor
$p=0.2$ of the box dimensions. The evidence margin is $\delta=0.03$: a claim
$c$ over region $x_i$ with label $\ell_i$ is retained if and only if
$f(x_i, c) \ge f(x_i, \ell_i) - \delta$, i.e., it must match the crop at least as
well as the bare object label up to the margin. Claims failing the gate fall
back to their object label, and multi-clause claims are salvaged clause-by-clause
before rejection. Geometric facts (counts, support, and spatial relations) are
derived deterministically from detection boxes and bypass the evidence gate. We
cap each image at $M_p$ person instances and $M_o$ object regions.

\begin{figure*}[t]
\centering
\begin{minipage}[c]{0.17\textwidth}
  \centering
  \includegraphics[width=\linewidth]{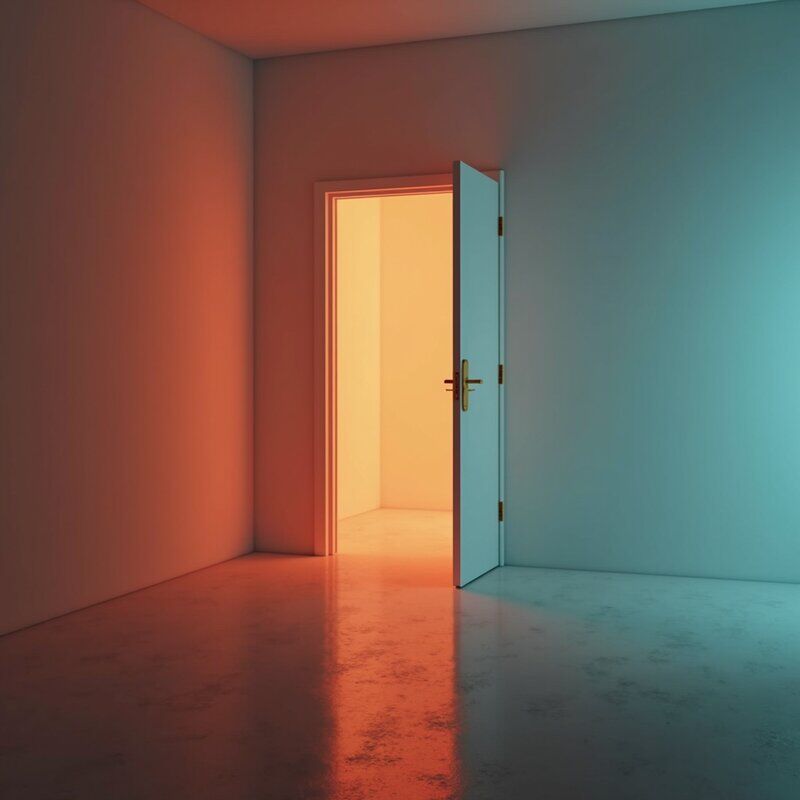}\\[4pt]
  \textbf{Target}
\end{minipage}\hfill
\begin{minipage}[c]{0.80\textwidth}
\renewcommand{\arraystretch}{1.3}
\setlength{\tabcolsep}{6pt}
\begin{tabular}{@{}>{\centering\arraybackslash}m{0.19\linewidth} m{0.75\linewidth}@{}}
\toprule
\textbf{Generated} & \textbf{Recovered prompt} \\
\midrule
\includegraphics[width=\linewidth]{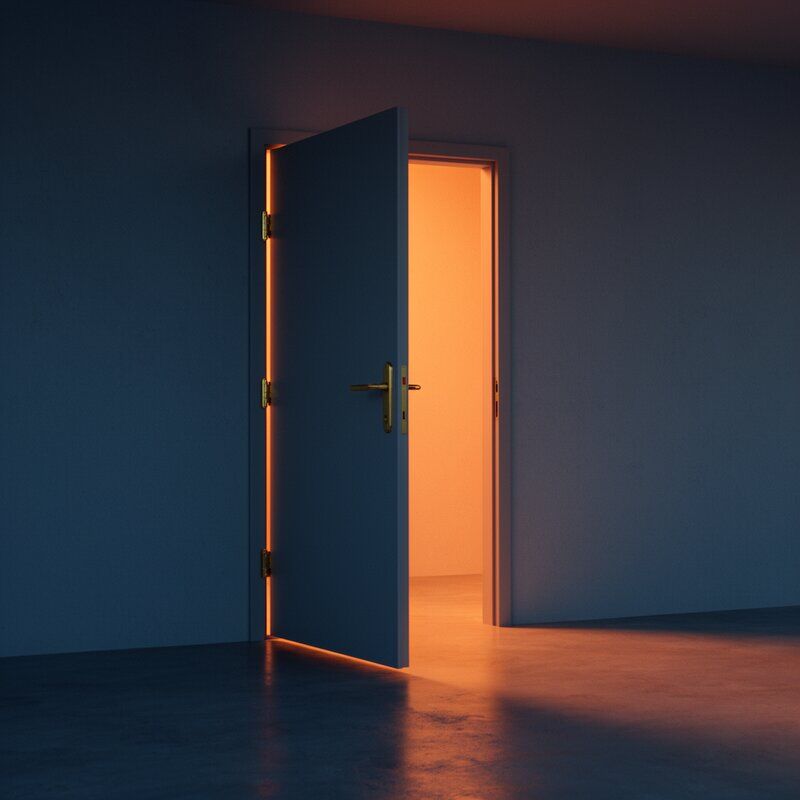} &
\textbf{Ours.} An open door emits warm light into a dimly lit room with cool blue and orange tones. The walls are plain, and the floor reflects the light, creating a stark contrast between the two halves of the space. Artificial, evening, indoor, eye-level, warm, orange, concrete floor, blurred. It contains three hinges, two walls. Gold handles, open, new, rectangular, smooth. No text, rectangular shape. Hinges. 3D render, minimalist, low-key, muted palette with orange accents, atmospheric, hyperdetailed. \\
\midrule
\includegraphics[width=\linewidth]{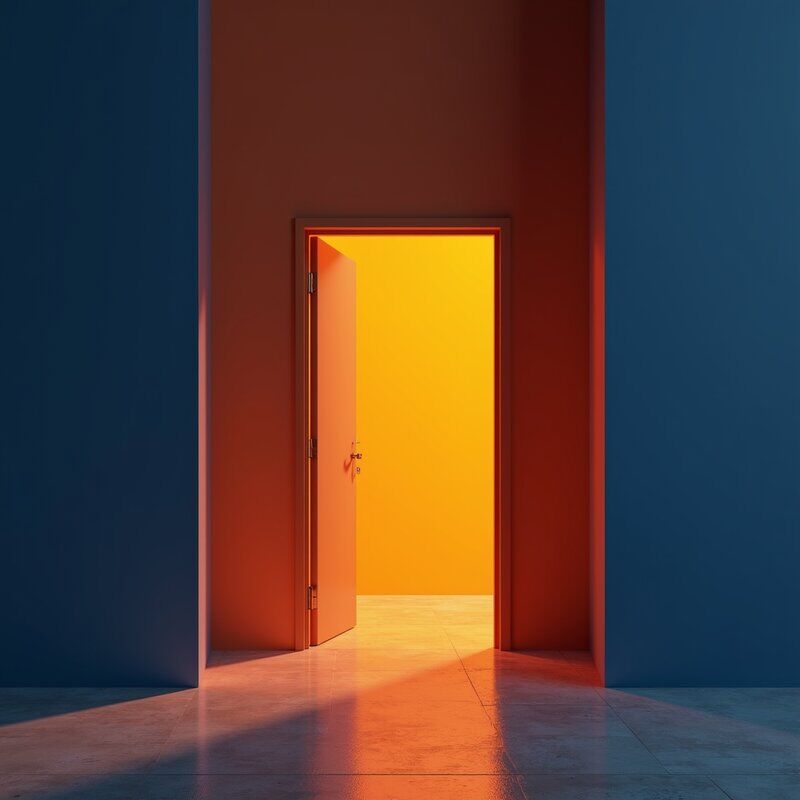} &
\textbf{PromptMiner.} an open door room, darker empty room with open doorway
leading to glowing lit hallway, warm lighting, orange and blue walls, concrete
floor, minimal composition, colorful, minimalist, high-contrast, high resolution. \\
\midrule
\includegraphics[width=\linewidth]{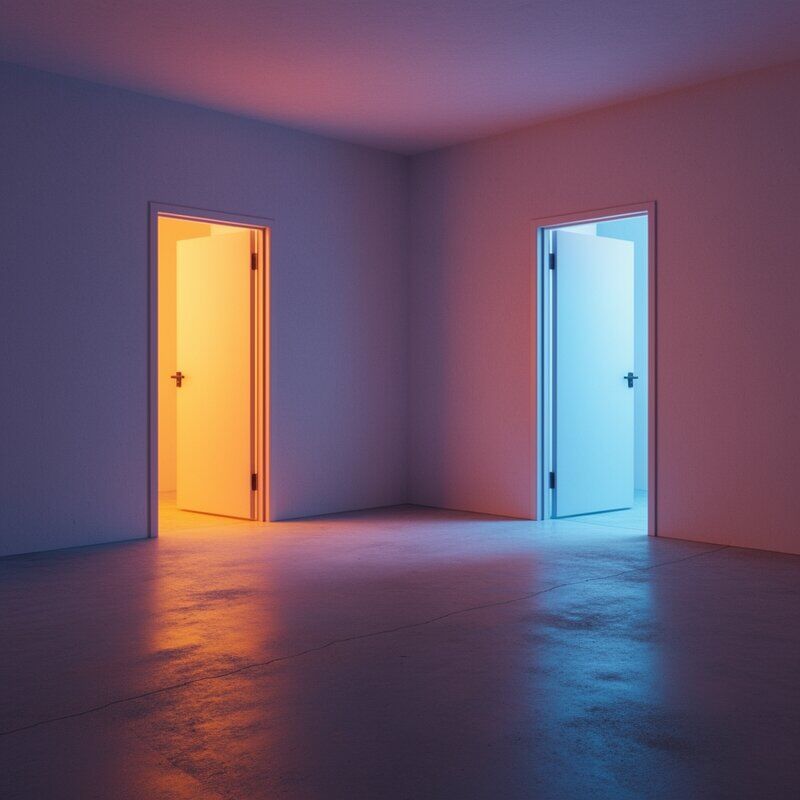} &
\textbf{VLM-as-expert.} A minimalist room with two open doors, one glowing warmly
and the other coolly, casting colorful reflections on the concrete floor. \\
\midrule
\includegraphics[width=\linewidth]{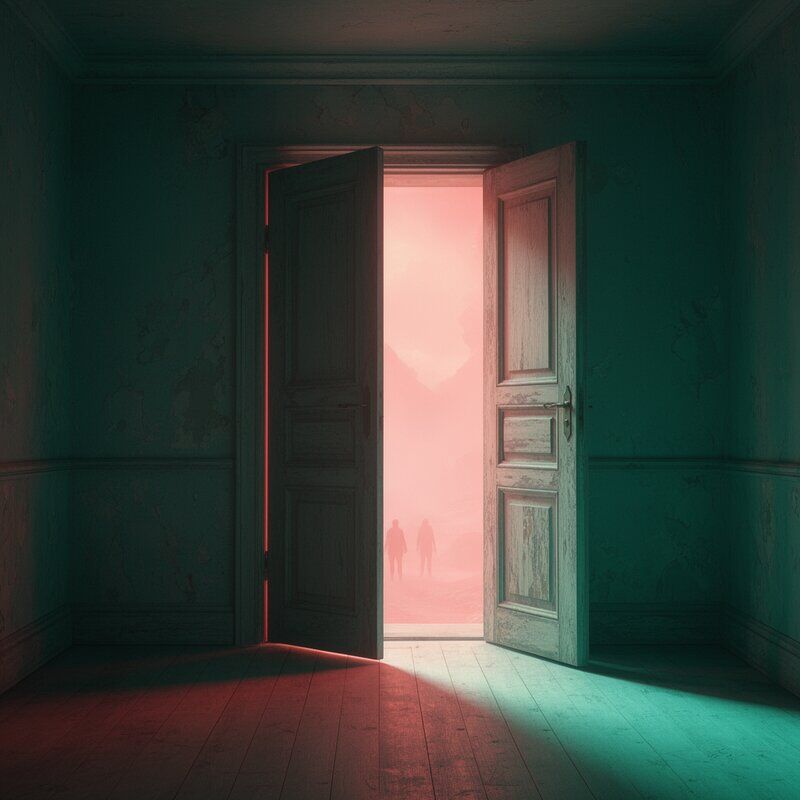} &
\textbf{CLIP-Interrogator.} a door in a room with a light coming in, teal orange
color palette 8k, pexels, high quality fantasy stock photo, low saturated red and
blue light, open door, vanilla-colored lighting, pink and teal and orange. \\
\bottomrule
\end{tabular}
\end{minipage}
\caption{Qualitative comparison on Lexica. We compare \emph{PROVE} against its three closest baselines, showing both the generated image and the recovered prompt. On these stylized images, \emph{PROVE} describes the visual style in grounded terms without attributing it to a named artist, unlike the baselines, which recover artist and platform tags.}
\label{fig:qual_lexica}
\end{figure*}

\paragraph{Generation and evaluation.}
Recovered prompts are rendered with FLUX.1-dev and SD3.5-Large under identical
sampling settings ($T$ inference steps, guidance scale $g$, fixed seeds
$\{s_1,s_2\}$) for all methods, so differences reflect the prompt rather than
the generator. We evaluate $100$ images each from MS-COCO, Flickr30K, and Lexica.

\paragraph{Hardware.}
All experiments run on a single NVIDIA L40S GPU. As \textsc{PROVE} is
inference-only, a full inversion requires one detector pass, one CLIP pass per
claim, and $O(\text{regions})$ VLM calls, with no backpropagation at any stage.

\end{document}